\documentclass[paper4A,10pt]{article}

\usepackage{color}
\usepackage{subfigure}
\usepackage{pifont}
\usepackage{amsmath,amssymb,amsthm,amsfonts,bm}
\usepackage{tikz}
\usepackage{3dplot}
\usetikzlibrary{shapes,arrows,shadows}
\usetikzlibrary{backgrounds}

\usepackage[latin1]{inputenc}
\usepackage{graphicx,float}
\usepackage{epsfig}
\usepackage{pdfpages}
\usepackage{psfrag}

\definecolor{tud9b}{rgb}{0.89,0.0,0.10}
\definecolor{tud8b}{rgb}{0.92,0.39,0.0}
\definecolor{tud2b}{rgb}{0.0,0.51,0.79}
\definecolor{tud3b}{rgb}{0.0,0.61,0.50}
\definecolor{tud4b}{rgb}{0.59,0.75,0.0}
\definecolor{tud6b}{rgb}{0.98,0.78,0.0}

\definecolor{tud0a}{rgb}{0.9,0.9,0.9}
\definecolor{tud0b}{rgb}{0.8,0.8,0.8}
\definecolor{tud0c}{rgb}{0.7,0.7,0.7}

\title{Some Aspects of Geometric Computer Vision for Analysing Dynamical Scenes focusing Automotive Applications}
\author{Volker Willert and Martin Buczko}

\begin{document}
\maketitle

\abstract{This draft summarizes some basics about geometric computer vision needed to implement efficient computer vision algorithms for applications that use measurements from at least one digital camera 
mounted on a moving platform with a special focus on automotive applications processing image streams taken from cameras mounted on a car. Our intention is twofold: On the one hand, we would like 
to introduce well-known basic geometric relations in a compact way that can also be found in lecture books about geometric computer vision like \cite{Ma2004, Hartley2003}. On the other hand, we would like to share
some experience about subtleties that should be taken into account in order to set up quite simple but robust and fast vision algorithms that are able to run in real time. 
We added a conglomeration of literature, we found to be relevant when implementing basic algorithms like optical flow, visual odometry and structure from motion.
The reader should get some feeling about how the estimates of these algorithms are interrelated, which parts of the algorithms are critical in terms of robustness and what kind of additional assumptions 
can be useful to constrain the solution space of the underlying usually non-convex optimization problems.}

\section{Introduction}

\begin{figure}
\begin{centering}
\includegraphics[width=0.75\linewidth]{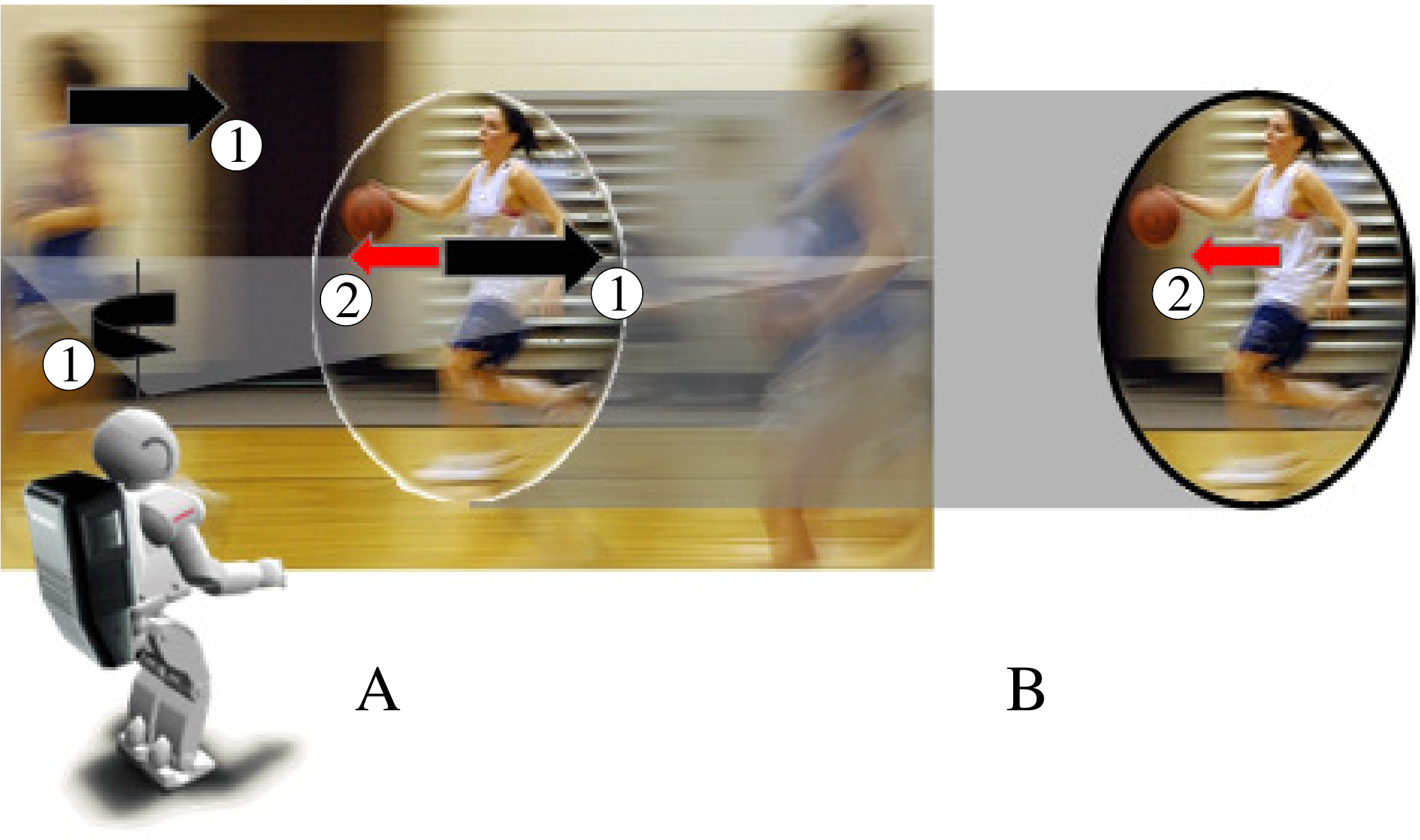}
\caption{The basic idea of consecutive motion estimation. Flow components that are induced by the observer \ding{172} and the observed objects \ding{173} are split up to isolate moving objects.}
\label{figMotiv}
\end{centering}
\end{figure}

To get an idea about what kind of estimation problems are relevant when analysing dynamical scenes, we sketch the task of moving object separation from visual observations of a moving 
platform starting with an illustrative example taken from \cite{Willert2008a}.
Assume, a mobile platform -- in this special case a mobile humanoid robot as can be seen in figure \ref{figMotiv} -- with a camera mounted on its head is moving around in a static
environment -- in this special case a gym -- while watching different moving objects -- in this special case the different players -- playing a basketball game. 

In case of ego-movements of the robot while observing moving objects, like turning the head left sketched in Fig.~\ref{figMotiv} A,
the visual flow is a superposition of projected movements of the objects \ding{173}, like the movement of the player running to the left and the egomotion induced 
flow field of the robot \ding{172}. Hence, the optical flow holds all the information describing the dynamics of the scene and
to separate moving objects from the background using images taken from a moving camera, a separation of the different
relative motions between the camera movement and the differently moving scene parts has to be realized. In this case, the movement of the player like depicted in Fig.~\ref{figMotiv} B
can only be separated if the relative movement of the robot to the background can be estimated properly. 
 
More precisely speaking, the movement of the robot in relation to the static background and the moving objects induces an optical flow field onto the image plane, 
which is a vector field consisting of vectors for each pixel of the image and every vector shows the projection of the movement of the correspondent world point projected onto the pixel.
So the optical flow field separates into different groups corresponding to different movements of projected rigid-object/background-parts \cite{Schmuedderich2008, Willert2009b} having some certain 3D structure.
Hence, there is a defined geometrical relation between the optical flow, the relative movement between the observer and the object/background-parts, and the 3D structure.
This relation is the essence to realize a separation of the moving objects from the static background using observations of a moving platform and was first formulated by Longuet-Higgins
et al. in 1980 \cite{Higgins1980}. Based on this famous geometrical constraint -- which today is called the continuous epipolar constraint \cite{Ma2004} -- one can solve for one of the three unknowns 
(flow, ego-motions, structure) once two unknowns can be estimated from suitable measurements and a segmentation of the different object parts is provided. 
Here, we are faced with a difficult dilemma: If the relative motions are not known and there is no other cue than optical flow 
to segment the parts that move differently, then both problems, the segmentation problem and the multiple motion estimation problem rely on each other \cite{Willert2006, Toussaint2007}. 

For the special case, if the robot moves around in a purely \textit{static} environment, then the estimation-segmentation dilemma does not arise. 
The projection of the environment onto the robot cameras induces a flow field that is \textit{exclusively} caused by the 
egomotion of the robot and varies with the 3D profile of the scene. Hence, there is only one relative motion and no segmentation is needed. 
Visual SLAM and egomotion computation approaches (also called visual odometry approaches) utilize these dependencies to 
estimate only the pose of a moving camera or the pose and in addition the scene structure usually assuming that a sparse and temporal stable set of point-to-point 
correspondences of static image features can be extracted. 

Additional sensing of the body movement via proprioception combined with the information of the visual flow allows for dense depth estimation which is called 
\textit{Structure from Motion}. As a reverse operation to egomotion-based depth estimation, the expected visual flow generated by egomotion can be inferred by 
combining body movement and scene depth information using depth cues like e.g. extracted from binocular disparity (\textit{Motion from Structure}). 

Unfortunately, in most cases the environment is \textit{not static} but contains moving objects. As already mentioned beforehand, 
these induce flow field components onto the robot cameras which deviate from the flow field as it is predicted from egomotion for static scenes.

Before delving into basic algorithms needed to realize applications like moving object detection from a moving camera system mounted on a car,
we provide the basic geometric relations.
We start with a mathematical description of the intertwining of optical flow, structure and ego-motion whereas a camera moves in a purely static scene. Then, 
we move on to a moving camera that moves in relation to a static background and several other moving objects. 

\section{Relation between Ego-motion, Optical Flow, and Structure of the Background}

The relation between ego-motion, optical flow, and structure can be formulated for the continuous and the discrete time case.
Historically, the continuous time case was formulated first \cite{Higgins1980}. Practically, the differetiation is relevant, because
the continuous motion case better models the case when the camera motion is slow compared to the camera frame rate and vice versa for
the discrete motion case. Formally, in the continuous motion case the twist of the camera motion is estimated whereas in the discrete case 
a pose change is estimated. 

\subsection{Continuous Rigid-Body-Transformations}

The 3D coordinates $\mathbf{X}_{\mathcal W} = \left[X_{\mathcal W}, Y_{\mathcal W}, Z_{\mathcal W}\right]^T$ of a fixed point $p_{\mathcal W} \in \mathbb{R}^3$ (in the background)
with respect to the world frame $\mathcal W$ and the 3D coordinates 
$\mathbf{X}_{\mathcal C} (t)= \left[X_{\mathcal C}(t), Y_{\mathcal C}(t), Z_{\mathcal C(t)}\right]^T$
of the same point with respect to a moving camera frame $\mathcal C$ are related by a rigid-body transformation $g=(\mathbf{R}_{\mathcal{CW}}(t), \mathbf{T}_{\mathcal{C}}(t)) \in SE(3)$ at time $t$  in the following way:
\begin{equation}\label{Eq1}
 \mathbf{X}_{\mathcal C}(t) = \mathbf{R}_{\mathcal{CW}}(t)\mathbf{X}_{\mathcal W} + \mathbf{T}_{\mathcal{C}}(t)\, ,
\end{equation}
with a rotational part denoted by a translation vector
$\mathbf{T}_{\mathcal{C}}(t) \in \mathbb{R}^3$ and a rotational part denoted by the rotation matrix $\mathbf{R}_{\mathcal{CW}}(t) \in SO(3)$.
This relation is also shown in figure \ref{figRigidMotionA}. The origin of the moving camera frame $\mathbf{X}_{\mathcal C}(t) = \mathbf{0}$ is given by
\begin{equation}
 \mathbf{0} = \mathbf{R}_{\mathcal{CW}}(t)\mathbf{T}_{\mathcal W}(t) + \mathbf{T}_{\mathcal{C}}(t) \quad \rightarrow \quad \mathbf{T}_{\mathcal{C}}(t) = -\mathbf{R}_{\mathcal{CW}}(t)\mathbf{T}_{\mathcal W}(t)\, ,
\end{equation}
and therefore the rigid-body transformation can also be formulated by the translation $\mathbf{T}_{\mathcal W}(t)$ of the camera with respect to the word frame as follows:
\begin{equation}
 \mathbf{X}_{\mathcal C}(t) = \mathbf{R}_{\mathcal{CW}}(t)\left(\mathbf{X}_{\mathcal W} - \mathbf{T}_{\mathcal{W}}(t)\right)\, ,
\end{equation}
also shown in figure \ref{figRigidMotionB}.

\begin{figure}
\centering
\parbox{2.0in}{
\tdplotsetmaincoords{65}{125}

\begin{tikzpicture}[scale=1,tdplot_main_coords]


\coordinate (P) at (-1,1.5,1);
\draw[color=black,-stealth,thick] (0,0,0) -- (P) node[color=black,midway, below, anchor=north west]{\scriptsize$\mathbf{X}_{\mathcal W}$};
\fill[color=black] (P) circle (0.05cm) node[anchor=south]{$p_{\mathcal W}$};

\begin{pgfonlayer}{background}
\draw[->] (0,0,0) -- (2.5,0,0) node[anchor=north, color=black]{\scriptsize $X_{\mathcal W}$};
\draw[->] (0,0,0) -- (0,2.5,0) node[anchor=south, color=black]{\scriptsize $Y_{\mathcal W}$};
\draw[->] (0,0,0) -- (0,0,2.5) node[anchor=south, color=black]{\scriptsize $Z_{\mathcal W}$};
\end{pgfonlayer}


\tdplotsetrotatedcoords{-36.20}{24.4}{53.79}
\coordinate (KK) at (1,-1,1.5) ;
\draw (KK) node[color=tud9b,anchor=east]{\scriptsize $\mathbf{R}_{\mathcal CW}(t)$};
\tdplotsetrotatedcoordsorigin{(KK)}
\draw[color=tud9b,tdplot_rotated_coords,->] (0,0,0) -- (0.5,0,0) node[anchor=north, color=tud9b]{\scriptsize $X_{\mathcal C}$};
\draw[color=tud9b,tdplot_rotated_coords,->] (0,0,0) -- (0,0.5,0) node[anchor=west, color=tud9b]{\scriptsize $Y_{\mathcal C}$};
\draw[color=tud9b,tdplot_rotated_coords,->] (0,0,0) -- (0,0,0.5) node[anchor=south, color=tud9b]{\scriptsize $Z_{\mathcal C}$};
\filldraw[color=tud0b,semitransparent,tdplot_rotated_coords] (1,-0.5,-0.5) -- (1,0.5,-0.5) -- (1,0.5,0.5) -- (1,-0.5,0.5);
\filldraw[color=tud0a,semitransparent,tdplot_rotated_coords] (1,-0.5,0.5) -- (1,0.5,0.5) -- (-1,0.5,0.5) -- (-1,-0.5,0.5);
\filldraw[color=tud0c,semitransparent,tdplot_rotated_coords] (1,0.5,0.5) -- (-1,0.5,0.5) -- (-1,0.5,-0.5) -- (1,0.5,-0.5);

\draw[color=tud9b,-stealth,thick] (KK) -- (P) node[color=tud9b,near end, above]{\scriptsize $\mathbf{X}_{\mathcal C}(t)$};
\draw[color=tud3b,-stealth,thick] (KK) -- (0,0,0) node[color=tud3b,midway, above, anchor=west]{\scriptsize $\mathbf{T}_{\mathcal C}(t)$};


\end{tikzpicture}
\caption{Rigid-body motion of a camera with respect to the \textbf{camera frame} (see direction of translation in cyan) \cite{Willert2012}.}
\label{figRigidMotionA}}%
\qquad
\begin{minipage}{2.0in}%
\tdplotsetmaincoords{65}{125}

\begin{tikzpicture}[scale=1,tdplot_main_coords]


\coordinate (P) at (-1,1.5,1);
\draw[color=black,-stealth,thick] (0,0,0) -- (P) node[color=black,midway, below, anchor=north west]{\scriptsize$\mathbf{X}_{\mathcal W}$};
\fill[color=black] (P) circle (0.05cm) node[anchor=south]{$p_{\mathcal W}$};

\begin{pgfonlayer}{background}
\draw[->] (0,0,0) -- (2.5,0,0) node[anchor=north, color=black]{\scriptsize $X_{\mathcal W}$};
\draw[->] (0,0,0) -- (0,2.5,0) node[anchor=south, color=black]{\scriptsize $Y_{\mathcal W}$};
\draw[->] (0,0,0) -- (0,0,2.5) node[anchor=south, color=black]{\scriptsize $Z_{\mathcal W}$};
\end{pgfonlayer}


\tdplotsetrotatedcoords{-36.20}{24.4}{53.79}
\coordinate (KK) at (1,-1,1.5) ;
\draw (KK) node[color=tud9b,anchor=east]{\scriptsize $\mathbf{R}_{\mathcal CW}(t)$};
\tdplotsetrotatedcoordsorigin{(KK)}
\draw[color=tud9b,tdplot_rotated_coords,->] (0,0,0) -- (0.5,0,0) node[anchor=north, color=tud9b]{\scriptsize $X_{\mathcal C}$};
\draw[color=tud9b,tdplot_rotated_coords,->] (0,0,0) -- (0,0.5,0) node[anchor=west, color=tud9b]{\scriptsize $Y_{\mathcal C}$};
\draw[color=tud9b,tdplot_rotated_coords,->] (0,0,0) -- (0,0,0.5) node[anchor=south, color=tud9b]{\scriptsize $Z_{\mathcal C}$};
\filldraw[color=tud0b,semitransparent,tdplot_rotated_coords] (1,-0.5,-0.5) -- (1,0.5,-0.5) -- (1,0.5,0.5) -- (1,-0.5,0.5);
\filldraw[color=tud0a,semitransparent,tdplot_rotated_coords] (1,-0.5,0.5) -- (1,0.5,0.5) -- (-1,0.5,0.5) -- (-1,-0.5,0.5);
\filldraw[color=tud0c,semitransparent,tdplot_rotated_coords] (1,0.5,0.5) -- (-1,0.5,0.5) -- (-1,0.5,-0.5) -- (1,0.5,-0.5);

\draw[color=tud9b,-stealth,thick] (KK) -- (P) node[color=tud9b,near end, above]{\scriptsize $\mathbf{X}_{\mathcal C}(t)$};
\draw[color=tud3b,-stealth,thick] (0,0,0) -- (KK) node[color=tud3b,midway, above, anchor=west]{\scriptsize $\mathbf{T}_{\mathcal W}(t)$};


\end{tikzpicture}
\caption{Rigid-body motion of a camera with respect to the \textbf{world frame} (see direction of translation in cyan) \cite{Willert2012}.}%
\label{figRigidMotionB}%
\end{minipage}%
\end{figure}%

If we now take the temporal derivative $\dot{\mathbf{X}}_{\mathcal C}(t)$ of the camera coordinates
\begin{equation}\label{Eq2}
 \dot{\mathbf{X}}_{\mathcal C}(t) = \dot{\mathbf{R}}_{\mathcal{CW}}(t)\mathbf{X}_{\mathcal W} + \dot{\mathbf{T}}_{\mathcal{C}}(t)\, ,
\end{equation}
solve \eqref{Eq1} for the point in world frame
\begin{equation}
 \mathbf{X}_{\mathcal W} = \mathbf{R}^T_{\mathcal{CW}}(t)\left(\mathbf{X}_{\mathcal C}(t) - \mathbf{T}_{\mathcal{C}}(t)\right)\, ,
\end{equation}
considering the inverse of the rotation matrix $\mathbf{R}^{-1}_{\mathcal{CW}}(t) = \mathbf{R}^T_{\mathcal{CW}}(t)$, and insert the result in \eqref{Eq2}, we arrive at the differential equation system
\begin{eqnarray}
 \dot{\mathbf{X}}_{\mathcal C}(t) & = & \dot{\mathbf{R}}_{\mathcal{CW}}(t)\mathbf{R}^T_{\mathcal{CW}}(t)\mathbf{X}_{\mathcal C}(t) + \dot{\mathbf{T}}_{\mathcal{C}}(t) - \mathbf{R}^T_{\mathcal{CW}}(t)\mathbf{T}_{\mathcal{C}}(t) \\
                                  & = & \hat{\bm{\omega}}(t)\mathbf{X}_{\mathcal C}(t) + \bm{\nu}(t)\, , \label{Eq3b} \\
 \left[\begin{array}{c} \dot{X}_{\mathcal C}(t) \\ \dot{Y}_{\mathcal C}(t) \\ \dot{Z}_{\mathcal C}(t)\end{array}\right] & = & 
 \left[\begin{array}{c} \omega_2(t)Z_{\mathcal C}(t)-\omega_3(t) Y_{\mathcal C}(t)+\nu_1(t) \\ \omega_3(t)X_{\mathcal C}(t)-\omega_1(t) Z_{\mathcal C}(t)+\nu_2(t) \\
 \omega_1(t)Y_{\mathcal C}(t)-\omega_2(t) X_{\mathcal C}(t)+\nu_3(t)
 \end{array}\right]\, ,
\end{eqnarray}
with the twist consisting of a rotational velocity $\bm{\omega}(t) = \left[\omega_1, \omega_2, \omega_3\right]^T$ 
and the corresponding skew-symmetric matrix $\hat{\bm{\omega}}(t) = \dot{\mathbf{R}}_{\mathcal{CW}}(t)\mathbf{R}^T_{\mathcal{CW}}(t) \in \mathbb{R}^{3\time 3}$, and
the translational velocity $\bm{\nu}(t) = \left[\nu_1, \nu_2, \nu_3\right]^T = \dot{\mathbf{T}}_{\mathcal{C}}(t) - \mathbf{R}^T_{\mathcal{CW}}(t)\mathbf{T}_{\mathcal{C}}(t)$.

\begin{figure}[h!]
\centering
\tdplotsetmaincoords{25}{150}

\begin{tikzpicture}[scale=1.5,tdplot_main_coords]


\draw[-stealth,dashed] (0,-3,0) -- (0,4.5,0) node[anchor=north]{\scriptsize optical axis};
\draw[color=tud8b,thick] (0,-2,0) -- (0,0,0) node[midway,above]{$f$};

\draw[->] (0,-2,0) -- (1,-2,0) node[anchor=east, color=black]{$Y_{\mathcal C}$};
\draw[->] (0,-2,0) -- (0,-1,0) node[anchor=east, color=black]{$Z_{\mathcal C}$};
\draw[->] (0,-2,0) -- (0,-2,1) node[anchor=south, color=black]{$X_{\mathcal C}$};

\tdplotsetrotatedcoords{0}{0}{120}
\coordinate (K) at (3,2.5,0) ;
\tdplotsetrotatedcoordsorigin{(K)}

\draw[->,tdplot_rotated_coords] (0,0,0) -- (1,0,0) node[anchor=west, color=black]{$Y_{\mathcal W}$};
\draw[->,tdplot_rotated_coords] (0,0,0) -- (0,1,0) node[anchor=south, color=black]{$Z_{\mathcal W}$};
\draw[->,tdplot_rotated_coords] (0,0,0) -- (0,0,1) node[anchor=south, color=black]{$X_{\mathcal W}$};
\draw[color=black] (2.8,2.7,0) node[anchor=west]{{\scriptsize world coordinates} $\mathbf{X}_{\mathcal W}$};

\draw[-stealth,color=tud9b] (3,2.5,0) -- (-1.25,3,1.25) node[midway,above, anchor=north west,color=black]{$\mathbf{X}_{\mathcal W}$};

\begin{pgfonlayer}{background}
\filldraw[color=tud0b,semitransparent] (1,-2,1) -- (-1,-2,1) -- (-1,-2,-1) -- (1,-2,-1);
\draw[color=tud0c] (-1,-2,1) node[anchor=west]{{\scriptsize camera coordinates} $\mathbf{X}$};
\end{pgfonlayer}
\fill[color=tud9b] (-1.25,3,1.25) circle (0.05cm) node[anchor=north]{$p$} node[color=black,anchor=south]{$\mathbf{X}$};
\fill[color=black] (0,-2,0) circle (0.05cm) node[anchor=west]{$o$};

\draw[-stealth,color=tud9b] (0,-2,0) -- (-1.25,3,1.25);


\draw[->] (0,0,0) -- (0.4,0,0) node[anchor=east, color=black]{$y$};
\draw[->] (0,0,0) -- (0,0,0.9) node[anchor=south, color=black]{$x$};
\begin{pgfonlayer}{background}
\filldraw[color=tud2b,semitransparent] (1.5,0,1.5) -- (-1.5,0,1.5) -- (-1.5,0,-1.5) -- (1.5,0,-1.5);
\draw[color=tud2b] (-1.5,0,1.5) node[anchor=west]{{\scriptsize image coordinates} $\mathbf{x}$};
\end{pgfonlayer}

\fill[color=tud9b] (-0.5,0,0.5) circle (0.05cm) node[anchor=south, color=black]{$\mathbf{x}$};
\draw[-stealth,color=tud9b] (0,0,0) -- (-0.5,0,0.5);



\draw[->] (-1,1.3,-1.5) -- (-1,1.3,-0.5);
\draw (-1,1.3,-0.75) node[anchor=west, color=black]{$x'$};
\draw[->] (-1,1.3,-1.5) -- (-0.5,1.3,-1.5);
\draw (-0.7,1.5,-1.7) node[color=black]{$y'$};
\fill[color=black] (0,1.3,0) circle (0.04cm) node[anchor=north]{$o_{\mathbf x}$};
\fill[color=tud9b] (-0.3,1.3,0) circle (0.05cm) node[anchor=south west, color=black]{$\mathbf{x}'$};
\draw[-stealth,color=tud9b] (-1,1.3,-1.5) -- (-0.3,1.3,0);

\begin{pgfonlayer}{background}
\filldraw[color=tud4b,semitransparent] (1,1.3,0.5) -- (-1,1.3,0.5) -- (-1,1.3,-1.5) -- (1,1.3,-1.5);
\draw[color=tud4b] (-1.1,1.3,0.8) node[anchor=west]{\scriptsize pixel};
\draw[color=tud4b] (-1.1,1.3,0.4) node[anchor=west]{{\scriptsize coordinates} $\mathbf{x}'$};
\draw[color=tud4b] (1,1.3,0.3) -- (-1,1.3,0.3);
\draw[color=tud4b] (1,1.3,0.1) -- (-1,1.3,0.1);
\draw[color=tud4b] (1,1.3,-0.1) -- (-1,1.3,-0.1);
\draw[color=tud4b] (1,1.3,-0.3) -- (-1,1.3,-0.3);
\draw[color=tud4b] (1,1.3,-0.5) -- (-1,1.3,-0.5);
\draw[color=tud4b] (1,1.3,-0.7) -- (-1,1.3,-0.7);
\draw[color=tud4b] (1,1.3,-0.9) -- (-1,1.3,-0.9);
\draw[color=tud4b] (1,1.3,-1.1) -- (-1,1.3,-1.1);
\draw[color=tud4b] (1,1.3,-1.3) -- (-1,1.3,-1.3);

\draw[color=tud4b](-1,1.3,-1.5) -- (1,1.3,-1.5);
\draw[color=tud4b](-0.8,1.3,0.5) -- (-0.8,1.3,-1.5);
\draw[color=tud4b](-0.6,1.3,0.5) -- (-0.6,1.3,-1.5);
\draw[color=tud4b](-0.4,1.3,0.5) -- (-0.4,1.3,-1.5);
\draw[color=tud4b](-0.2,1.3,0.5) -- (-0.2,1.3,-1.5);
\draw[color=tud4b](0,1.3,0.5) -- (0,1.3,-1.5);
\draw[color=tud4b](0.2,1.3,0.5) -- (0.2,1.3,-1.5);
\draw[color=tud4b](0.4,1.3,0.5) -- (0.4,1.3,-1.5);
\draw[color=tud4b](0.6,1.3,0.5) -- (0.6,1.3,-1.5);
\draw[color=tud4b](0.8,1.3,0.5) -- (0.8,1.3,-1.5);

\end{pgfonlayer}

\end{tikzpicture}
\caption{Perspective projection from 3D world coordinates to 2D pixel coordinates of a calibrated camera with known intrinsic parameters \cite{Willert2012}.}%
\label{figProjection}%
\end{figure}
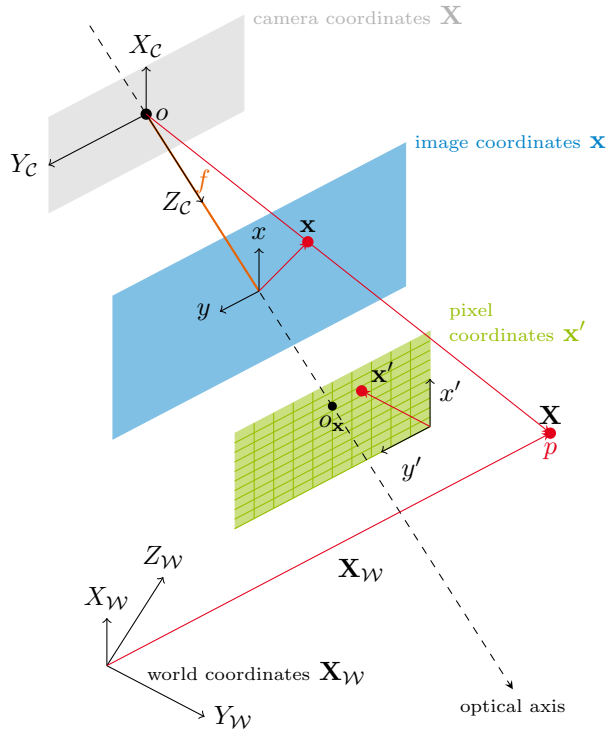

Practically, equation \eqref{Eq3b} is relevant if the camera motion is slow compared to the camera frame rate. So, for low-speed car applications in inner-city scenarios
with for example stop-and-go movements this continuous rigid-body-transformation is relevant. One has to keep in mind, that equation \eqref{Eq3b} is a purely kinematic model
and the real dynamics (like taking care of the mass of the car) is neglected. Also, changes in camera acceleration are not taken into account. Nevertheless, this pure kinematic
model is accurate enough because inertia of the car is large, so acceleration changes are very small in practice.

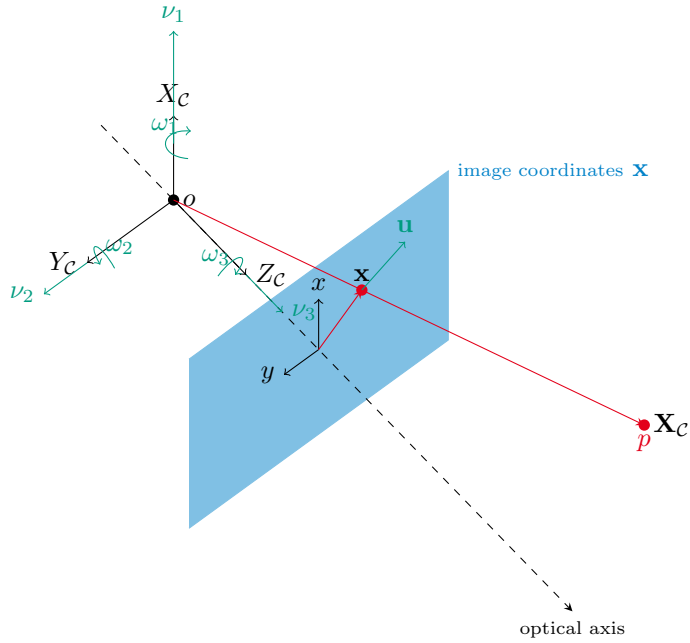
\begin{figure}[h!]
\centering
\tdplotsetmaincoords{30}{140}

\begin{tikzpicture}[scale=1.5,tdplot_main_coords]


\draw[-stealth,dashed] (0,-3,0) -- (0,3.5,0) node[anchor=north]{\scriptsize optical axis};

\draw[->] (0,-2,0) -- (1,-2,0) node[anchor=east, color=black]{$Y_{\mathcal C}$};
\draw[->] (0,-2,0) -- (0,-1,0) node[anchor=west, color=black]{$Z_{\mathcal C}$};
\draw[->] (0,-2,0) -- (0,-2,1.5) node[anchor=south, color=black]{$X_{\mathcal C}$};

\draw[color=tud3b,->] (0.85,-1.8,0) .. controls (0.85,-2,0.2) and (0.85,-2.2,0).. node[anchor=west]{$\omega_2$} (0.85,-2,-0.2);
\draw[color=tud3b,->] (0.2,-1.15,0) .. controls (0,-1.15,0.2) and (-0.2,-1.15,0).. node[anchor=east]{$\omega_3$} (0,-1.15,-0.2);
\draw[color=tud3b,->] (0,-1.8,1) .. controls (0.2,-2,1) and (0,-2.2,1).. node[anchor=south]{$\omega_1$} (-0.2,-2,1);

\draw[color=tud3b,->] (0,-2,1) -- (0,-2,3) node[anchor=south]{$\nu_1$};
\draw[color=tud3b,->] (1,-2,0) -- (1.5,-2,0) node[anchor=east]{$\nu_2$};
\draw[color=tud3b,->] (0,-1,0) -- (0,-0.5,0) node[anchor=west]{$\nu_3$};




\fill[color=tud9b] (-1.25,3,1.25) circle (0.05cm) node[anchor=north]{$p$} node[color=black,anchor=west]{$\mathbf{X}_{\mathcal C}$};
\fill[color=black] (0,-2,0) circle (0.05cm) node[anchor=west]{$o$};

\draw[-stealth,color=tud9b] (0,-2,0) -- (-1.25,3,1.25);


\draw[->] (0,0,0) -- (0.4,0,0) node[anchor=east, color=black]{$y$};
\draw[->] (0,0,0) -- (0,0,0.9) node[anchor=south, color=black]{$x$};
\begin{pgfonlayer}{background}
\filldraw[color=tud2b,semitransparent] (1.5,0,1.5) -- (-1.5,0,1.5) -- (-1.5,0,-1.5) -- (1.5,0,-1.5);
\draw[color=tud2b] (-1.5,0,1.5) node[anchor=west]{{\scriptsize image coordinates} $\mathbf{x}$};
\end{pgfonlayer}

\fill[color=tud9b] (-0.5,0,0.5) circle (0.05cm) node[anchor=south, color=black]{$\mathbf{x}$};
\draw[-stealth,color=tud9b] (0,0,0) -- (-0.5,0,0.5);
\draw[color=tud3b,->] (-0.5,0,0.5) -- (-1.0,0,0.8) node[anchor=south]{$\mathbf{u}$};

\end{tikzpicture}
\caption{Optical flow $\mathbf{u}$ at image coordinate $\mathbf{x}$ induced by a continuous camera motion $(\bm{\omega},\bm\nu)$ and a fixed scene point $p$ with coordinates $\mathbf{X}_{\mathcal C}$.}%
\label{figContMotion}%
\end{figure}

Now, the perspective projection
\begin{equation}
 \lambda \overline{\mathbf{x}}' = \bm{\Pi} \overline{\mathbf{X}}_{\mathcal W}\, ,
\label{Eq9}
\end{equation}
with $\lambda$ being an arbitrary scaling factor and the projection matrix 
\begin{equation}
 \bm{\Pi} = \mathbf{K} \left[\begin{array}{c} \mathbf{R}_{\mathcal CW} | \mathbf{T}_{\mathcal C} \end{array}\right] \in \mathbb{R}^{3 \times 4}\, ,
\end{equation}
has to be taken into account to derive the relation between image coordinates $\mathbf{x}(t) = \left[x(t), y(t)\right]^T$, 
the optical flow $\mathbf{u}(t) = \left[u(t), v(t)\right]^T$, and the structure/depth $Z_\mathcal{C}(t)$.
The general projection is shown in figure \ref{figProjection} all the way from 3D coordinates in world frame written in homogeneous coordinates 
$\overline{\mathbf{X}}_{\mathcal W} = \left[X_{\mathcal W}, Y_{\mathcal W}, Z_{\mathcal W}, 1\right]^T$, 
along the transformation to the camera frame in homogeneous coordinates $\overline{\mathbf{X}}_{\mathcal C} = \left[X_{\mathcal C}, Y_{\mathcal C}, Z_{\mathcal C}, 1\right]^T$, then the projection
to the image frame in homogeneous coordinates $\overline{\mathbf{x}} = \left[x, y, 1\right]^T$, and finally the transformation to the pixel frame 
$\overline{\mathbf{x}}' = \left[x', y', 1\right]^T$.
If we assume a calibrated camera and the kalibration matrix
\begin{equation}
   \mathbf{K}=\left[\begin{array}{ccc} fs_x & fs_\theta & o_x \\ 0 & fs_y & o_y \\ 0 & 0 & 1 \end{array}\right]\, , 
\end{equation} 
with the focal length $f$, the principal point $\mathbf{o} = \left[o_x, o_y\right]^T$, and the scaling factors $s_x, s_y, s_\theta$
to be known (these are the intrinsic parameters of the camera), then we can always transform from pixel coordinates to the normalized image coordinates as follows
\begin{equation}
 \overline{\mathbf{x}}=\mathbf{K}^{-1}\overline{\mathbf{x}}'\,.
\end{equation}
Neglecting the time index for brevity, we can now derive the relation between image coordinates $\mathbf{x} = \left[x, y\right]^T$, 
the optical flow $\mathbf{u} = \left[u, v\right]^T$, and the structure/depth $Z_\mathcal{C}$ using the normalized projection
\begin{equation}\label{Eq5b}
   \mathbf{x} = \left[\begin{array}{c} x \\ y \end{array}\right]=\frac{1}{Z_{\mathcal C}}\left[\begin{array}{c} X_{\mathcal C} \\ Y_{\mathcal C} \end{array}\right] \quad \longleftrightarrow \quad Z_{\mathcal C} \overline{\mathbf{x}} = \mathbf{X}_{\mathcal C}\,, 
\end{equation}
and its temporal derivative
\begin{equation}\label{Eq6}
 \dot{\mathbf{X}}_{\mathcal C} = \dot{Z}_{\mathcal C} \overline{\mathbf{x}} + Z_{\mathcal C}\dot{\overline{\mathbf{x}}} \, .
\end{equation}
Inserting equation \eqref{Eq6} into equation \eqref{Eq3b}, we obtain
\begin{equation}\label{Eq7}
 \dot{\overline{\mathbf{x}}} = \left[\begin{array}{c} u \\ v \\ 0 \end{array}\right] = \hat{\bm{\omega}}\overline{\mathbf{x}} + \frac{1}{Z_{\mathcal C}} \bm{\nu} - \frac{\dot{Z}_{\mathcal C}}{Z_{\mathcal C}}\overline{\mathbf{x}} \, .
\end{equation}
Solving the third row for the temporal derivative of the depth $\dot{Z}_{\mathcal C} = \omega_1 y Z_{\mathcal C}-\omega_2 x Z_{\mathcal C} + \nu_3$ and inserting the result into the first two rows of matrix equation \eqref{Eq7},
we get the basic relation between optical flow, structure, and ego-motion:
\begin{eqnarray}
 u & = & \underbrace{(\nu_1-x\nu_3)/Z_{\mathcal C}}_{u_T} + \underbrace{\omega_2(1+x^2)-\omega_3 y - \omega_1 xy}_{u_R}\, , \nonumber \\
 v & = & \underbrace{(\nu_2-y\nu_3)/Z_{\mathcal C}}_{v_T} + \underbrace{\omega_1(1-y^2)+\omega_3 x + \omega_2 xy}_{v_R}\, . \label{Eq8}
\end{eqnarray}
The relations are shown in figure \ref{figContMotion}. The optical flow $\mathbf{u} = \mathbf{u}_T + \mathbf{u}_R$ is therefore the vector sum of a translational component $\left[u_T, v_T\right]^T$ that is independent of the rotational velocity $\bm{\omega}$ 
but dependent on the depth of the projected 3D points $Z_{\mathcal C}$ and a rotational component $\left[u_R, v_R\right]^T$ that is independent of the 3D structure of the scene but dependent on second 
order monomials of the image coordinates $(xy, x^2, y^2)$. This knowledge is very important to tune visual odometry algorithms, because it clearly shows the different levels of sensitivity to measurement
noise in the image coordinates and the optical flow of the different ego-motion parameters \cite{Buczko2016}. It also clearly proves, that the rotational velocity should be estimated first \cite{Buczko2016b}, because no structure information is needed
and thus, also no error propagation from noisy depth measurements can enter. Finally, it proves, that monocular odometry approaches \cite{Buczko2017} can only estimate the rotational velocity and the translational velocity
up to an unknown scale. So the scale is purely dependent on the scene structure and if no scene structure can be measured via for example a stereo camera system, then additional 
prior assumptions about the scene structure or additional sensors have to be taken into account to estimate this scale.

Equation \eqref{Eq8} shows, that if the depth $Z_{\mathcal C}$ is not known, then only the ratios $\nu_1/Z_{\mathcal C},\nu_2/Z_{\mathcal C},\nu_3/Z_{\mathcal C}$ or in other words
only the direction of the translational velocity vector can be estimated. Thus, the unknown scale can be eliminated from equation \eqref{Eq7} considering the inner product of the vectors in 
equation \eqref{Eq7} with the vector $(\bm{\nu} \times \overline{\mathbf{x}})$. This leads to the continuous epipolar constraint
\begin{eqnarray}
 & \dot{\overline{\mathbf{x}}}^T\widehat{\bm{\nu}}\overline{\mathbf{x}} = \overline{\mathbf{x}}^T\widehat{\bm{\omega}}^T\widehat{\bm{\nu}}\overline{\mathbf{x}} & \, , \\
 & \overline{\mathbf{u}}^T\widehat{\bm{\nu}}\overline{\mathbf{x}} + \overline{\mathbf{x}}^T\widehat{\bm{\omega}}\widehat{\bm{\nu}}\overline{\mathbf{x}}  =  0 & \, . \label{Eq9a}
\end{eqnarray}

\subsection{Discrete Rigid-Body-Transformations}

If the camera motion is fast compared to the camera frame rate a discrete rigid-body-transformation can be used instead
\begin{equation}\label{Eq4}
 \mathbf{X}_{\mathcal C}(t+\Delta t) = \Delta\mathbf{R}\mathbf{X}_{\mathcal C}(t) + \Delta\mathbf{T}\, ,
\end{equation}
with the time difference $\Delta t$, the relative translation $\Delta\mathbf{T}$, and the relative rotation $\Delta\mathbf{R}$ between two camera frames 
$\mathcal{C}$ at two time instants $t$ and $t+\Delta t$. 

Practically, equation \eqref {Eq4} better models high-speed car-applications in outer-city/highway scenarios. Equation \eqref{Eq4} assumes a constant twist of the camera motion
between two different consecutive camera poses, Thus, it is violated in case of large camera accelerations.

If you now plug equation \eqref{Eq5b} for two different time instants 
\begin{eqnarray}
  \mathbf{X}_{\mathcal C}(t) & = & Z_{\mathcal C}(t) \overline{\mathbf{x}}(t)\, , \\
  \mathbf{X}_{\mathcal C}(t+\Delta t) & = & Z_{\mathcal C}(t+\Delta t) \overline{\mathbf{x}}(t+\Delta t)\, ,
\end{eqnarray}
into equation \eqref{Eq4} you arrive at the discrete relation between image coordinate displacements $\Delta \mathbf{x} = \mathbf{x}(t+\Delta t)-\mathbf{x}(t)$,
the depths $Z_{\mathcal C}(t)$ and $Z_{\mathcal C}(t+\Delta t)$ and the discrete camera motion $(\Delta\mathbf{R},\Delta\mathbf{T})$ which reads
\begin{eqnarray}\label{Eq22}
 Z_{\mathcal C}(t+\Delta t) \overline{\mathbf{x}}(t+\Delta t) & = & Z_{\mathcal C}(t)\Delta\mathbf{R}\overline{\mathbf{x}}(t) + \Delta\mathbf{T}\, ,\\
\overline{\mathbf{x}}(t) + \Delta \overline{\mathbf{x}} & = & \frac{Z_{\mathcal C}(t)}{Z_{\mathcal C}(t+\Delta t)}\Delta\mathbf{R}\overline{\mathbf{x}}(t) + \frac{\Delta\mathbf{T}}{Z_{\mathcal C}(t+\Delta t)} \, , \\
 \Delta \overline{\mathbf{x}} & = & \left[\frac{Z_{\mathcal C}(t)}{Z_{\mathcal C}(t\!+\!\Delta t)}\Delta\mathbf{R} \! - \! \mathbf{I}\right]\overline{\mathbf{x}}(t) \! + \! \frac{\Delta\mathbf{T}}{Z_{\mathcal C}(t\!+\!\Delta t)} . \label{Eq10}
\end{eqnarray}
This equation is the discrete equivalent to equation \eqref{Eq7}. As in the continuous case the depths can be eliminated and the discrete motion can be estimated up to a scale
with a monocular camera solving the discrete epipolar constraint which reads
\begin{equation}\label{Eq26}
 \overline{\mathbf{x}}^T(t+\Delta t)\widehat{\Delta\mathbf{T}}\Delta\mathbf{R}\overline{\mathbf{x}}(t) = 0\, .
\end{equation}


\subsection{The Discrete versus the Continuous Epipolar Constrained}

There do exist several well known algorithms to solve equation \eqref{Eq9a} and \eqref{Eq26} for the ego-motion up to the unknown scale of translation, namely the discrete and continuous 
8-point- \cite{Ma2004} and the five-point-algorithm \cite{Nister2004} with different kinds of realisations, like \cite{Hartley2012}.
Here, it is common sense within the community and confirmed on lots of test data and different visual odometry configurations 
that the five-point-algorithm outperforms the eight-point-algorithm \cite{Scaramuzza2011, Fraundorfer2012}.
Two important aspects have to be mentioned. On the one hand, since the five-point-algorithm provides an analytical solution that solves a nonlinear equation system using the minimal number of points
the system designer has to take care of chosing the right five points to base the estimate on (e.g. using a RANSAC algorithm). 
Thus, this kind of solution outperforms linear least squares approaches, like the eight-point-algorithm in terms of accuracy. On the other hand, in terms of robustness, the least squares approaches
are more robust. Sometimes, the five-point-algorithm fails because of bad feature configurations. 

One conclusion from this knowledge could be: In order to bring together both advantages nonlinear least squares approaches could be a step
towards very precise and robust solutions also for fairly bad feature configurations. This could be one direction to be evaluated more thoroughly.

It is very important for the algorithms to work properly that the points used for solving the equations are in \textit{general position}. If the points are on so called 
\textit{critical surfaces} the algorithms fail to solve uniquely for the ego-motion because several solutions do exist. 
A case of practical importance occurs when all the points happen to lie on the same plane. 
This has to be known in advance and can be solved uniquely  with a reduced four-point-algorithm.

There is another requirement for the algorithms to work properly which is \textit{sufficient parallax}. This means, the translation between two frames cannot be zero and should 
have some certain absolute value. Otherwise, the translation estimate is wrong but the rotation estimate is usually correct.

The discrete case assumes a distingtion of the vantage points from two consecutive views. If this is not the case only the continuous motion case works.
Further on, the \textit{twisted-pair ambiguity} that occurs in the discrete case and can be solved by imposing a positive depth constraint does not appear 
in the continuous motion case. Further details on the algorithms can be found in \cite{Ma2004}.

\subsection{Structure reconstruction without known scale}
\label{SecScale}

If only a mono camera is used the structure can only be reconstructed up to an unknown scale.
This is important for monocular visual odometry approaches that want to add a local mapping for improving motion estimates. 
Once the pose change $(\gamma\Delta\mathbf{T},\Delta\mathbf{R})$ has been estimated up to the single universal unknown scale $\gamma$
the depth's for all points $j$ can be recovered up to this scale ambiguity using equation \eqref{Eq22}:
\begin{equation}
 Z_{\mathcal C}(t+\Delta t,j) \overline{\mathbf{x}}(t+\Delta t,j)  =  Z_{\mathcal C}(t,j)\Delta\mathbf{R}\overline{\mathbf{x}}(t,j) + \gamma\Delta\mathbf{T}\,.
\end{equation}
Since the depth with respect to the current frame $Z_{\mathcal C}(t,j)$ is redundant to the depth with respect to the next frame $Z_{\mathcal C}(t+\Delta t,j)$,
one depth can be eliminated
\begin{eqnarray}
 Z_{\mathcal C}(t,j) \widehat{\overline{\mathbf{x}}}(t+\Delta t,j)\Delta\mathbf{R}\overline{\mathbf{x}}(t,j) +  \gamma\widehat{\overline{\mathbf{x}}}(t+\Delta t,j)\Delta\mathbf{T}  & = & 0\,, \\
 \left[\widehat{\overline{\mathbf{x}}}(t+\Delta t,j)\Delta\mathbf{R}\overline{\mathbf{x}}(t,j), 
       \widehat{\overline{\mathbf{x}}}(t+\Delta t,j)\Delta\mathbf{T}\right]\left[Z_{\mathcal C}(t,j), \gamma\right]^T & = & 0\,.
\end{eqnarray}
Stacking all these linear equations for all points $j$ into one matrix leads to a linear least squares estimate for all depth's $Z_{\mathcal C}(t,j)\,, \forall j$ up to the unknown scale $\gamma$.
Details can be found in \cite{Ma2004}.

\subsection{The Reprojection Error}

Up to now, we considered an ideal pinhole camera model with a calibrated camera and ideal point correspondences $\mathbf{x}(t,j) \rightarrow \mathbf{x}(t+\Delta t,j)$.
In practice, these coordinates are noisy $\tilde{\mathbf{x}}(t,j)$ because of the limited resolution of the camera chip and the ambiguities in the correspondence search. In addition, the
camera cannot be calibrated such that the ideal pinhole model holds very precisely. 
Finally, the correspondences are obtained with an optical flow algorithm that usually does not consider the epipolar constraint.
Hence, the measured (noisy) coordinates do not satisfy the epipolar constraint precisely.

To account for these errors both the pose estimate and the coordinate correspondences can be refined using the reprojection error which is the euklidean
distance between the measured and reprojected coordinates from last and current time frame subject to the epipolar constraint:
\begin{equation}
 \epsilon(t,j) = |\!| \tilde{\mathbf{x}}(t-\Delta t,j) -\pi(\Delta\mathbf{R}Z_{\mathcal C}(t,j)\mathbf{x}(t,j)+\Delta\mathbf{T})|\!|_2\,.
\label{Eq29}
\end{equation}
Here $\pi$ denotes the projection given in \eqref{Eq5b}.
The reprojection error $\epsilon(t,j)$ is heavily used to formulate non-convex optimization problems if a precise ego-motion and structure
reconstruction is needed.   

\subsection{Bundle Adjustment}
\label{SecBundle}

In order to use the reprojection error for ego-motion and structure refinement several solutions do exist that are summarized with \textit{bundle adjustment} (BA). 
The most general one is the full bundle adjustment approach that optimizes some likelihood function $\theta$ of the reprojection errors at different times $t$ of a bundle
of points $j$. In order to find the global optimum of the non-convex objective for several ego-motions along several frames within a time interval 
the selection of so called keyframes to get a good initialization is a crucial point. If the initialization is too bad such that the inital guess of the ego-motions is too far away from
the optimum, the frames cover bad configurations like insufficient parallax and there is no sufficient overlap between the points of the bundle projected to the different frames then the
iterative gradient-based optimization schemes do not converge to a proper solution \cite{Thormaehlen2004}. 
 
For visual odometry applications a local bundle adjustment approach (using only a small number of timely consecutive frames) 
is more suitable which in the simplest case boils down to a two-frame bundle adjustment for each frame pair.
If the objective should not account for outliers then the sum of squared reprojection errors is a proper objective function and one arrives at the classical least squares estimator
chosing the current frame as the reference frame:
\begin{equation}
  \theta(\Delta\mathbf{R}, \Delta\mathbf{T}, \mathbf{x}(t,j), Z_{\mathcal C}(t,j)) = \sum\limits_j |\!|\tilde{\mathbf{x}}(t,j)-\mathbf{x}(t,j)|\!|_2^2 + (\epsilon(t,j))^2\,.
\end{equation}
The optimization of this objective drastically reduces the drift in visual odometry applications but needs some extra time for computation.
Since it is solved iteratively based on a gradient descent approach the choice of the gradient method and which parameters to optimize at which iteration
influences the final result. Thus, a lot of different approaches are around with difference in accuracy, convergence speed and computational cost. 
Optimizing for the ego-motion only $\theta(\Delta\mathbf{R}, \Delta\mathbf{T})$ assuming the depth's and 2D coordinates are given, which means the structure is not refined
and the errors because of calibration are negligible, is called \textit{motion-only} BA. Optimizing for the depth's and 2D coordinates only
$\theta(\mathbf{x}(t,j), Z_{\mathcal C}(t,j))$, thus refining the 3D points given a suitable ego-motion estimate, is called \textit{structure-only} BA or \textit{structure triangulation}.
Recent approaches try to find alternating optimization schemes that switch between \textit{motion-only}, \textit{structure-only} and \textit{full-local} BA to reach extremely precise
ego-motion estimates with only a small amount of additional computational cost \cite{Forster2014}.

\section{Optical Flow}

Optical flow is one of the main ingredients to visual odometry, moving object detection and motion analysis approaches.
Thus, it is very important to rely on a robust and computationally efficient optical flow estimation technique.

\subsection{Photometric Constraint -- The Brightness Constancy Assumption}

The basic equation included in all optical flow estimation techniques is a photometric constraint also called the
brightness constancy assumption. It assumes that the brightness $I(\mathbf{x},t):= I_1(\mathbf{x})$ of a 3D point projected onto an image plane of a camera
and measured by the camera chip at position $\mathbf{x}$ at time $t$ does not change during the movement of the point in space. The resulting projection
after some time $\Delta t$ and pixel movement $\mathbf{u}$ should have the same brightness $I(\mathbf{x}+\mathbf{u},t+\Delta t):= I_2(\mathbf{x}+\mathbf{u})$.
Here, the pixel movement is the forward movement from frame $1$ to frame $2$ in accordance to the Lucas-Kanade formulation.
Thus, the brightness constancy assumption can be formulated as follows:
\begin{equation}
 I(\mathbf{x},t):= I_1(\mathbf{x}) \approx I_2(\mathbf{x}+\mathbf{u}) =: I(\mathbf{x}+\mathbf{u},t+\Delta t)\,.
\label{Eq31}
\end{equation}
 
\subsection{Lucas-Kanade Optical Flow}

Lucas and Kanade extended this assumption to some neighborhood $\mathcal{N}_\mathbf{x}$ around each pixel and assumed
that the movement of the projection of this neighborhood (patch) is constraint by an image warp $\mathcal{T}(\mathbf{x};\mathbf{p})$
whereas the warp defines a local optical flow field parameterized with $\mathbf{p}$. For the warp a lot of different models with different
complexity and number of parameters can be defined, e.g. a homography, an affine warp, or simply a translation.
Also the way how to parameterize the same warp can be different. Now you have some additional contraint on the movement of the
pixels within some neighborhood and you can formulate an objective function based on the photometric constraint \eqref{Eq31}.
The one proposed by Lucas-Kanade is the most prominent one and simply realizes a sum of least squares objective:
\begin{equation}
 J(\mathbf{p}) := \sum\limits_{\mathbf{x} \in \mathcal{N}_\mathbf{x}} \left[I_2(\mathcal{T}(\mathbf{x};\mathbf{p}))-I_1(\mathbf{x})\right]^2\,.
\label{Eq32}
\end{equation}
This is a nonlinear optimization problem even if we have a warp function that is linear in $\mathbf{p}$
because the pixel values $I(\mathbf{x})$ are in general non-linear in $\mathbf{x}$ or in fact totally un-related.
The optimization problem to find the optimal motion parameters $\hat{\mathbf{p}}$ can be formulated as follows:
\begin{equation}
 \hat{\mathbf{p}} = \mbox{argmin}_\mathbf{p} J(\mathbf{p})\,.
\end{equation}
Such nonlinear optimization problems are solved iteratively via a proper linearization. In order to find a good
local minimum a suitable initialization $\mathbf{p}_0$ has to be given and each iteration $i$ solves for an increment
of these parameters such that   
\begin{equation}
 \Delta \mathbf{p}_i = \mbox{argmin}_{\Delta \mathbf{p}} J(\mathbf{p}_i+\Delta \mathbf{p})\,, \quad \mbox{with update} \quad \mathbf{p}_{i+1} = \mathbf{p}_i + \Delta \mathbf{p}_i\, .
\label{Eq34}
\end{equation}
The iteration is repeated until the norm of the increment $|\!|\Delta \mathbf{p}_i|\!| < \epsilon$ is below some threshold $\epsilon$ or a maximum number of iterations is reached.
In order to solve the optimization \eqref{Eq34} a first order Taylor expansion on the second image is applied
\begin{equation}
 I_2(\mathcal{T}(\mathbf{x};\mathbf{p}+\Delta \mathbf{p})) \approx I_2(\mathcal{T}(\mathbf{x};\mathbf{p})) + \nabla I_2(\mathcal{T}(\mathbf{x};\mathbf{p})) \frac{\partial\mathcal{T}(\mathbf{x};\mathbf{p})}{\partial\mathbf{p}}\Delta \mathbf{p}\,.
\label{Eq35}
\end{equation}
Inserting this approximation into the optimization problem \eqref{Eq34} leads to a closed form solution:
\begin{equation}
 \Delta \mathbf{p} = \left[\sum\limits_{\mathbf{x} \in \mathcal{N}_\mathbf{x}} \left[\nabla I_2 \frac{\partial\mathcal{T}}{\partial\mathbf{p}}\right]^T\left[\nabla I_2 \frac{\partial\mathcal{T}}{\partial\mathbf{p}}\right]\right]^{-1}\sum\limits_{\mathbf{x} \in \mathcal{N}_\mathbf{x}}\left[\nabla I_2 \frac{\partial\mathcal{T}}{\partial\mathbf{p}}\right]^T\left[I_1(\mathbf{x})-I_2(\mathcal{T}(\mathbf{x};\mathbf{p}))\right]\,.
\label{Eq36}
\end{equation}
For the most simplest warp $\mathcal{T}(\mathbf{x};\mathbf{p})=[x+u, y+v]^T$ which is the same shift $\mathbf{p}=[u,v]^T$ for all pixels $\mathbf{x}=[x,y]^T$ in the neighborhood
$\mathcal{N}_\mathbf{x}$ the Jacobian of the warp equals the identity $\frac{\partial\mathcal{T}(\mathbf{x};\mathbf{p})}{\partial\mathbf{p}} = \mathbf{I}$ and 
the equations reduce to the well-known Lucas-Kanade optical flow equations:
\begin{equation}
 \Delta \mathbf{u} = \underbrace{\left[\sum\limits_{\mathbf{x} \in \mathcal{N}_\mathbf{x}} \left[\left[\nabla I_2 \right]^T\left[\nabla I_2 \right]\right]\right]^{-1}\sum\limits_{\mathbf{x} \in \mathcal{N}_\mathbf{x}}\left[\nabla I_2 \right]^T}_{\mbox{re-compute at each iteration}}\left[I_1(\mathbf{x})-I_2(\mathbf{x}+\mathbf{u})\right]\,.
\end{equation}
Usually, the neighborhood is weighted using a Gaussian window to relax the Lucas-Kanade constraint with distance to the center point of the neighborhood.
  
\subsection{Why to use Lucas-Kanade?} The local and linear differential method of Lucas and
Kanade \cite{Kanade1981} is one of the most popular approaches for optical flow computation when it comes to real time applications with restricted computational power. 
There are several reasons for that. 
Compared to global smoothness constraints used for example by Horn and Schunk, their local 
explicit method is more accurate and more robust with respect to errors in gradient measurements.
It is very easy to compute and real-time capable.  
Nevertheless, the local approach suffers from the aperture problem and the linearisation of the 
underlying constancy assumption for image intensity. Lucas' and Kanade's basic idea to
assume that the optical flow field is spatially constraint within some neighborhood is in many cases 
not enough to resolve motion ambiguities and does in particular not hold at motion
boundaries. Further on, the linearised intensity constancy assumption is suitable only for small displacements.

However, if only a sparse flow is needed and a proper feature detector is used in advance to 
operate only on patches that have unambiguous structure, like corners, which is especially true for
all sparse visual odometry approaches, the Lukas-Kanade approach is still one of the most efficient
and robust optical flow methods if used in combination with a pyramidal approach \cite{Bouguet2001}.

\subsection{Efficient Implementation of Lucas-Kanade}

The implementation of the pyramidal approach is straight forward and details about implementation can be found for example in \cite{Bouguet2001}
that describes an efficient implementation done by Intel Corporation.

A more interesting detail on implementation is the way how the warp should be applied along the iterations within a scale.
Here, a computationally efficient method is proposed by Baker and Matthews, called the \textit{inverse compositional} algorithm
\cite{Baker2004}. Since this is relevant for real-time applications the basic idea is outlined in the following.
In equation \eqref{Eq36}, it is important to mention that the gradient $\nabla I_2(\mathcal{T}(\mathbf{x};\mathbf{p}))$ must be evaluated at
$\mathcal{T}(\mathbf{x};\mathbf{p})$ and the Jacobian $\frac{\partial\mathcal{T}(\mathbf{x};\mathbf{p})}{\partial\mathbf{p}}$ must be evaluated
at $\mathbf{p}$. So, both depend in general on $\mathbf{p}$.
Thus, in general both have to be re-calculated at each interation step because both depend on $\mathbf{p}$. For simple motion models, like linear motion models,
the Jacobian is constant and needs not to be re-calculated.
In equation \eqref{Eq36}, the computation of the Hessian $ \mathbf{H} = \sum\limits_{\mathbf{x} \in \mathcal{N}_\mathbf{x}}\left[\nabla I_2 \frac{\partial\mathcal{T}}{\partial\mathbf{p}}\right]^T\left[\nabla I_2 \frac{\partial\mathcal{T}}{\partial\mathbf{p}}\right]$
is the computationally most expensive step with computational complexity of $\mathcal{O}(n^2N)$. Here, $n$ is the number of motion parameters and $N$ the number of pixels.

So, the idea of the \textit{inverse compositional} algorithm \cite{Baker2004} is to reformulate the Lucas-Kanade approach such that the Hessian needs not to be re-calculated for each iteration $i$.
The \textit{compositional} approach is an alternative way to optimize equation \eqref{Eq34} but is completely equivalent. First optimize the following objective for $\Delta \mathbf{p}$
\begin{equation}
 \Delta \mathbf{p}_i = \mbox{argmin}_{\Delta \mathbf{p}} \sum\limits_{\mathbf{x} \in \mathcal{N}_\mathbf{x}} \left[I_2(\mathcal{T}(\mathbf{x};\mathbf{p}_i) \circ \mathcal{T}(\mathbf{x};\Delta\mathbf{p}))-I_1(\mathbf{x})\right]^2\,,
\end{equation}
and then update the warp 
\begin{equation}
 \mathcal{T}(\mathbf{x};\mathbf{p}_{i+1}) = \mathcal{T}(\mathbf{x};\mathbf{p}_i) \circ \mathcal{T}(\mathbf{x};\Delta\mathbf{p}_i)\,.
\end{equation}
The \textit{inverse compositional} approach now reverses $I_1$ and $I_2$ and inverts the warp of the motion increment:  
\begin{equation}
 \Delta \mathbf{p}_i = \mbox{argmin}_{\Delta \mathbf{p}} \sum\limits_{\mathbf{x} \in \mathcal{N}_\mathbf{x}} \left[I_1(\mathcal{T}(\mathbf{x};\Delta\mathbf{p}))-I_2(\mathcal{T}(\mathbf{x};\mathbf{p}_i))\right]^2\,,
\label{Eq40}
\end{equation}
and then updates the warp
\begin{equation}
 \mathcal{T}(\mathbf{x};\mathbf{p}_{i+1}) = \mathcal{T}(\mathbf{x};\mathbf{p}_i) \circ \mathcal{T}(\mathbf{x};\Delta\mathbf{p}_i)^{-1}\,.
\label{Eq41}
\end{equation}
Following this equivalence the optimization via Taylor expansion reads:
\begin{equation}
 I_1(\mathcal{T}(\mathbf{x};\Delta \mathbf{p})) \approx I_1(\mathcal{T}(\mathbf{x};\mathbf{0})) + \nabla I_1(\mathbf{x}) \frac{\partial\mathcal{T}(\mathbf{x};\mathbf{0})}{\partial\mathbf{p}}\Delta \mathbf{p}\,.
\end{equation}
Inserting this approximation into the optimization problem \eqref{Eq40} leads to a closed form solution:
\begin{equation}
 \Delta \mathbf{p} = \left[\sum\limits_{\mathbf{x} \in \mathcal{N}_\mathbf{x}} \left[\nabla I_1 \frac{\partial\mathcal{T}}{\partial\mathbf{p}}\right]^T\left[\nabla I_1 \frac{\partial\mathcal{T}}{\partial\mathbf{p}}\right]\right]^{-1}\sum\limits_{\mathbf{x} \in \mathcal{N}_\mathbf{x}}\left[\nabla I_1 \frac{\partial\mathcal{T}}{\partial\mathbf{p}}\right]^T\left[I_2(\mathcal{T}(\mathbf{x};\mathbf{p}))-I_1(\mathbf{x})\right]\,.
\end{equation}
For the most simplest warp $\mathcal{T}(\mathbf{x};\mathbf{p})=[x+u, y+v]^T$ which is the same shift $\mathbf{p}=[u,v]^T$ for all pixels $\mathbf{x}=[x,y]^T$ in the neighborhood
$\mathcal{N}_\mathbf{x}$ the Jacobian of the warp equals the identity $\frac{\partial\mathcal{T}(\mathbf{x};\mathbf{0})}{\partial\mathbf{p}} = \mathbf{I}$ and 
the Hessian $\sum\limits_{\mathbf{x} \in \mathcal{N}_\mathbf{x}} \left[\nabla I_1 \right]^T\left[\nabla I_1 \right]$ does not depend on $\mathbf{p}$ anymore.
Thus, the Hessian needs to be computed only once and is constant along iterations $i$. This reduces the computational complexity because a large part 
of the iterative update can be pre-computed:
\begin{equation}
 \Delta \mathbf{u} = \underbrace{\left[\sum\limits_{\mathbf{x} \in \mathcal{N}_\mathbf{x}} \left[\nabla I_1 \right]^T\left[\nabla I_1 \right]\right]^{-1}\sum\limits_{\mathbf{x} \in \mathcal{N}_\mathbf{x}}\left[\nabla I_1 \right]^T}_{\mbox{pre-compute only once}}\left[I_2(\mathbf{x}+\mathbf{u})-I_1(\mathbf{x})\right]\,.
\end{equation}
Inverting and composing of the warps in equation \eqref{Eq41} need some small extra costs $\mathcal{O}(n^2)$ that is almost negligible.
So, if more than one iteration is applied, the inverse compositional approach should be considered. If a pyramidal approach is chosen, then usually only one iteration at each
scale is enough because the result of the current scale is used as an initialization for the next finer scale and thus needs not to be very precise. 
Only at the scale with highest resolution several iterations could be beneficial because this scale
deliveres the final result with highest precision. 
 
\subsection{Extensions to Lucas-Kanade}

The basic idea of Lucas and Kanade is to constrain the local motion measurement by assuming a constant velocity
within a spatial neighborhood. In \cite{Willert2007, Willert2008b} this spatial constraint is reformulated 
in a probabilistic way assuming Gaussian distributed uncertainty in spatial identification of velocity
measurements and extended to scale and time dimensions. Thus, uncertain velocity
measurements observed at different image scales and positions over time can be combined.  
This leads to a recurrent optical flow filter formulated in a Dynamic Bayesian Network applying
suitable factorisation assumptions and approximate inference techniques. 
The introduction of spatial uncertainty allows for a dynamic and spatially adaptive tuning of 
the constraining neighborhood. The tuning is realized dependent on the local structure tensor 
of the intensity patterns of the image sequence. It is demonstrated that a probabilistic
combination of spatiotemporal integration and modulation of a purely local integration area improves the Lucas and
Kanade estimation. 
This idea is further extended in \cite{Willert2005, Willert2009a} to deal with any kind of distribution providing 
a general belief propagation scheme. 

Other attempts to extend and improve the Lukas-Kanade approach are second order approaches and robust norms \cite{Baker2003}.
Unfortunately, those extensions do only lead to marginal improvement or even worsen the results but requiring a much higher computational cost.

\section{Prior Models for Automotive Applications}

Prior models are models that include knowledge that is known a priori, that is before running an algorithm, taking measurements
and making observations. In the car domain that includes kowledge about the kinematics and dynamics of the ego-car \cite{Buczko2018} 
or a car in general if one observes also other traffic participants \cite{Schreier2014} and knowledge about the geometry of the scene, 
like the surface \cite{Schreier2012, Schreier2013, Schreier2015}. Beside prior models, there is also prior information about the scenario for example from digital maps and other categories 
like inner city or highway, property of the ground and weather conditions.
Additional to the prior models and informations there are helpful quantities, like the calibration parameters including the hight over ground
of the camera.

Prior information can be used to adapt the algorithms to the scenario using most suitable parameters or switch between different algorithms solving the same task under different conditions.
Prior models are more powerful and can help to reduce the parameter space and thus lead to more efficient and robust algorithms but less accuracy because
the model does not cover all effects of the real physical world in all situations.

\subsection{Prior Scene Models}
\label{SecStructure}

The most often used scene model is moving on a planar ground. There are only a few extensions to the plane, like curved surfaces.

\subsubsection{The planar ground assumption}

\begin{figure}
\centering
\tdplotsetmaincoords{45}{150}

\begin{tikzpicture}[scale=1,tdplot_main_coords]


\draw[->] (0,-2,0) -- (1,-2,0) node[anchor=east, color=black]{$Y_{\mathcal{C}}$};
\draw[->] (0,-2,0) -- (0,-1,0) node[anchor=east, color=black]{$Z_{\mathcal{C}}$};
\draw[->] (0,-2,0) -- (0,-2,1) node[anchor=south, color=black]{$X_{\mathcal{C}}$};

\begin{pgfonlayer}{background}
\filldraw[color=tud0b,semitransparent] (1,-2,1) -- (-1,-2,1) -- (-1,-2,-1) -- (1,-2,-1);
\draw[color=tud0c] (-1,-2,1) node[anchor=west]{{\scriptsize camera coordinates} $\mathbf{X}_{\mathcal{C}}$};
\end{pgfonlayer}

\fill[color=black] (0,-2,0) circle (0.05cm) node[anchor=south west]{$o$};


\draw[->] (0,0,0) -- (0.4,0,0) node[anchor=east, color=black]{$y$};
\draw[->] (0,0,0) -- (0,0,0.9) node[anchor=south, color=black]{$x$};
\begin{pgfonlayer}{background}
\filldraw[color=tud2b,semitransparent] (1.5,0,1.5) -- (-1.5,0,1.5) -- (-1.5,0,-1.5) -- (1.5,0,-1.5);
\draw[color=tud2b] (-1.5,0,1.5) node[anchor=west]{{\scriptsize image coordinates} $\mathbf{x}$};
\end{pgfonlayer}

\filldraw[color=tud0b,semitransparent] (1.5,0.5,-1.5) -- (-1.5,0.5,-1.5) -- (-1.5,3.5,-1.5)   -- (1.5,3.5,-1.5);

\draw [color=tud6b] (-1.5,0.5,-1.5) -- (-1.5,3.5,-1.5);
\draw [color=tud6b] (1.5,0.5,-1.5) -- (1.5,3.5,-1.5);
\draw [color=tud6b] (-1.2,0,-1.2)  -- (-0.55,0,-0.55);
\draw [color=tud6b] (1.2,0,-1.2) -- (0.55,0,-0.55);

\filldraw[color=tud0b,semitransparent] (1.2,0,-1.2) -- (-1.2,0,-1.2)  -- (-0.55,0,-0.55) -- (0.55,0,-0.55);
\draw [color=tud6b] (1.5,0.5,-1.5) -- (-1.5,0.5,-1.5);
\draw [color=tud6b] (1.5,1,-1.5) -- (-1.5,1,-1.5);
\draw [color=tud6b] (1.5,1.5,-1.5) -- (-1.5,1.5,-1.5);
\draw [color=tud6b] (1.5,2,-1.5) -- (-1.5,2,-1.5);
\draw [color=tud6b] (1.5,2.5,-1.5) -- (-1.5,2.5,-1.5);
\draw [color=tud6b] (1.5,3,-1.5) -- (-1.5,3,-1.5);
\draw [color=tud6b] (1.5,3.5,-1.5) -- (-1.5,3.5,-1.5);

\draw [color=tud6b] (1.2,0,-1.2) -- (-1.2,0,-1.2);
\draw [color=tud6b] (1,0,-1) -- (-1,0,-1);
\draw [color=tud6b] (0.86,0,-0.86) -- (-0.86,0,-0.86);
\draw [color=tud6b] (0.75,0,-0.75) -- (-0.75,0,-0.75);
\draw [color=tud6b] (0.66,0,-0.66) -- (-0.66,0,-0.66);
\draw [color=tud6b] (0.6,0,-0.6) -- (-0.6,0,-0.6);
\draw [color=tud6b] (0.55,0,-0.55) -- (-0.55,0,-0.55);

\draw[-stealth,color=tud9b] (0,-2,0) -- (1.5,0.5,-1.5);
\fill[color=tud9b] (1.5,0.5,-1.5) circle (0.05cm);
\fill[color=tud9b] (1.2,0,-1.2) circle (0.05cm);
\draw[-stealth,color=tud9b] (0,-2,0) -- (-1.5,0.5,-1.5);
\fill[color=tud9b] (-1.5,0.5,-1.5) circle (0.05cm);
\fill[color=tud9b] (-1.2,0,-1.2) circle (0.05cm);

\draw[-stealth,color=tud9b] (-0.55,0,-0.55) -- (-1.5,3.5,-1.5);
\fill[color=tud9b] (-1.5,3.5,-1.5) circle (0.05cm);
\fill[color=tud9b] (-0.55,0,-0.55) circle (0.05cm);
\draw[-stealth,color=tud9b] (0.55,0,-0.55) -- (1.5,3.5,-1.5);
\fill[color=tud9b] (1.5,3.5,-1.5) circle (0.05cm);
\fill[color=tud9b] (0.55,0,-0.55) circle (0.05cm);

\draw[-stealth,dashed] (0,-3,0) -- (0,4.5,0) node[anchor=north]{\scriptsize optical axis};

\end{tikzpicture}
\caption{Planar ground assumption and homography.}%
\label{figHomography}%
\end{figure}
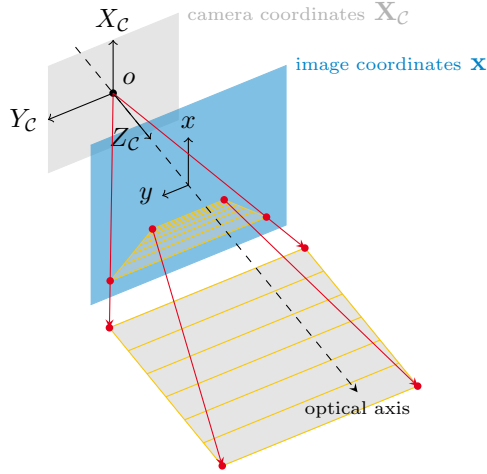

In the automotive domain a lot of approaches include the planar ground assumption (e.g. MobileEye and many others).
Here, the free space to drive on (e.g. the road) is assumed to be a plane. Figure \ref{figHomography} shows the dependencies of a camera with
a planar ground (grey) projected onto the image plane (blue). This special kind of projection is called a homography. 
A plane with $\mathbf{n}=[n_1,n_2,n_3]^T$ and $|\!|\mathbf{n}|\!|=1$ being the surface normal and $d$ being the distance from the optical center to the plane 
in camera coordinates is given by equation $\mathbf{n}^T\mathbf{X}_{\mathcal C} = d$.
Solving for $Z_{\mathcal C}$ and inserting into the projective equation \eqref{Eq9} leads to a homography:
\begin{eqnarray}
   Z_{\mathcal C} \overline{\mathbf{x}} & = & \left[\begin{array}{ccc} \mid & \mid & \mid \\
                              (\mathbf{\pi}_1-\frac{n_1}{n_3}\mathbf{\pi}_3) &
                              (\mathbf{\pi}_2-\frac{n_2}{n_3}\mathbf{\pi}_3) &
                              (\mathbf{\pi}_4-\frac{d}{n_3}\mathbf{\pi}_3) \\
                              \mid & \mid & \mid\end{array} \right]
                         \left[\begin{array}{c} X_\mathcal{C} \\
                                                Y_\mathcal{C} \\
                                                1 \end{array} \right] \\
                                        & = & \mathbf{H}\left[\begin{array}{c} X_\mathcal{C} \\
                                                Y_\mathcal{C} \\
                                                1 \end{array}\right]\,,
\end{eqnarray}
whereas $\mathbf{\pi}_i$ are the columns of the canonical projection matrix \cite{Ma2004}.  

\subsubsection{Extended ground ssumptions and Free Space Detection}

\begin{figure}
\begin{centering}
\includegraphics[width=0.9\linewidth]{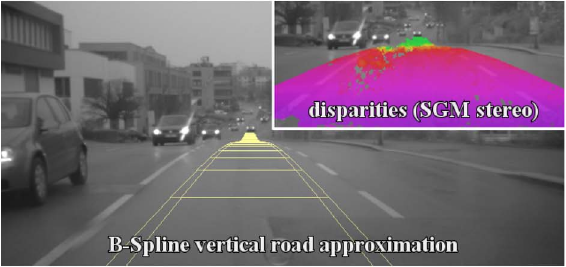}
\caption{Example of a B-spline fit along the longitudinal direction taken from \cite{Wedel2009}.}
\label{figBSplineRoad}
\end{centering}
\end{figure}

\begin{figure}
\begin{centering}
\includegraphics[width=0.9\linewidth]{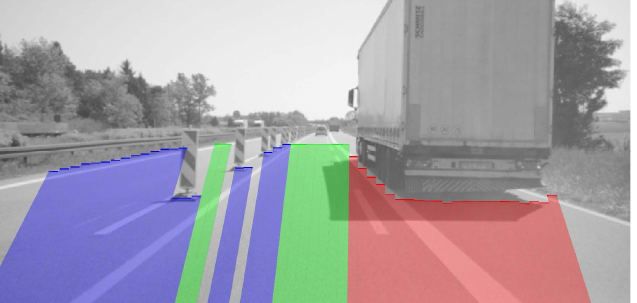}
\caption{An example of a free space detection taken from \cite{Neumann2015}.}
\label{figFreeSpace}
\end{centering}
\end{figure}

There are some few extensions to the ground plane assumption. Especially in situations with slope changes in the road course ahead due to appraoching a hill or a dip 
cannot be modelled apropriately with a planar model. One interesting approach as can be seen in figure \ref{figBSplineRoad} is a B-spline extension fitted into 
the longitudinal structure of the surface extracted from a V-disparity map and stabilized via a Kalman-filter of the B-spline parameters along frames \cite{Wedel2009}.
The most general representation of drivable space is the so called free space detection. One example of a free space detection is shown in figure \ref{figFreeSpace}.
A nice review on current approaches can be found in \cite{Neumann2015}.

\subsection{Prior Motion Models}
\label{SecMotion}

There are several suitable approximations for the camera motion mounted on a car including
more or less explicit knowledge about the motion constraints of the camera due to inertia of the car (e.g. slow motions)
and a constraint motion set because of the ability of a car to move (e.g. single-track-model).

\begin{figure}[h!]
\centering
\tdplotsetmaincoords{30}{140}

\begin{tikzpicture}[scale=1.5,tdplot_main_coords]


\draw[-stealth,dashed] (0,-3,0) -- (0,3.5,0) node[anchor=north]{\scriptsize optical axis};

\draw[->] (0,-2,0) -- (1,-2,0) node[anchor=east, color=black]{$Y_{\mathcal C}$};
\draw[->] (0,-2,0) -- (0,-1,0) node[anchor=west, color=black]{$Z_{\mathcal C}$};
\draw[->] (0,-2,0) -- (0,-2,1.5) node[anchor=south, color=black]{$X_{\mathcal C}$};

\draw[color=tud3b,->] (0.85,-1.8,0) .. controls (0.85,-2,0.2) and (0.85,-2.2,0).. node[anchor=west]{$\omega_2$} (0.85,-2,-0.2);
\draw[color=tud3b,->] (0.2,-1.15,0) .. controls (0,-1.15,0.2) and (-0.2,-1.15,0).. node[anchor=east]{$\omega_3$} (0,-1.15,-0.2);
\draw[color=tud3b,->] (0,-1.8,1) .. controls (0.2,-2,1) and (0,-2.2,1).. node[anchor=south]{$\omega_1$} (-0.2,-2,1);

\draw[color=tud3b,->] (0,-2,1) -- (0,-2,3) node[anchor=south]{$\nu_1$};
\draw[color=tud3b,->] (1,-2,0) -- (1.5,-2,0) node[anchor=east]{$\nu_2$};
\draw[color=tud3b,->] (0,-1,0) -- (0,-0.5,0) node[anchor=west]{$\nu_3$};




\begin{pgfonlayer}{background}
\filldraw[color=tud2b,semitransparent] (1.5,0.5,1.5) -- (-1.5,0.5,1.5) -- (-1.5,0.5,-1.5) -- (1.5,0.5,-1.5);
\draw[color=tud2b] (-1.5,0,1.5) node[anchor=west]{{\scriptsize image plane}};
\end{pgfonlayer}


\end{tikzpicture}
\caption{Camera movements consisting of three translational components, $\nu_1$, $\nu_2$ and $\nu_3$ for the velocities along the $X$-, $Y$-, and $Z$-axis respectively and
the rotational velocity components, namely yaw $\omega_1$, pitch $\omega_2$, and roll $\omega_3$.}%
\label{figContMotion2}%
\end{figure}
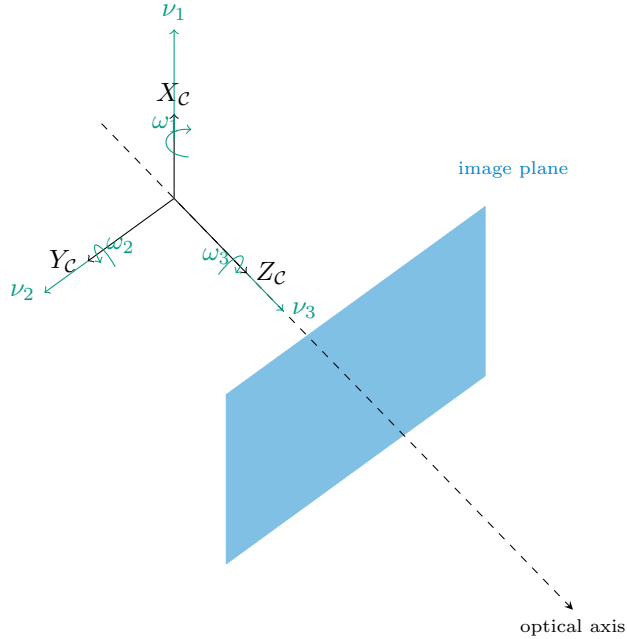

Repeating the general motion equations and the relation to optical flow and structure \eqref{Eq8}
\begin{eqnarray}
 u & = & (\nu_1-x\nu_3)/Z_{\mathcal C} + \omega_2(1+x^2)-\omega_3 y - \omega_1 xy\, , \nonumber \\
 v & = & (\nu_2-y\nu_3)/Z_{\mathcal C} + \omega_1(1-y^2)+\omega_3 x + \omega_2 xy\, , \nonumber
\end{eqnarray}
we have three translational components, $\nu_1$, $\nu_2$ and $\nu_3$ for the velocities along the $X$-, $Y$-, and $Z$-axis respectively and
the rotational velocity components, namely yaw $\omega_1$, pitch $\omega_2$, and roll $\omega_3$, shown again in figure \ref{figContMotion2}.

\subsubsection{The small rotation approximation}

A suitable approximation if rotations are small is the following: $\Delta\mathbf{R} \approx \mathbf{I}+\widehat{\bm{\omega}}$.
This can be used if translational motion components are dominating and no turns are driven. It could be interesting if only an
instantaneous motion estimate is needed to separate moving objects from the background at high speeds of a car moving on a highway.
In addition, this is a suitable approximation in post-processing recursive refinement schemes that are based on optimizing an objective
to refine the motion estimates incrementally (e.g. using the reprojection error).

\subsubsection{The three parameter motion model}

Especially for outlier rejection it is advantageous to reduce the general motion model to a minimum \cite{Stein2000, Fraundorfer2012}.
The motion of a car can be modeled as a translation $\nu_3$ along the $Z_{\mathcal C}$ axis and rotation $\omega_1, \omega_2$ around the
$X_{\mathcal C}$ and $Y_{\mathcal C}$ axes. Limiting the equations \eqref{Eq8} to this motion constraints, we get:
\begin{eqnarray}
 u & = & -x\nu_3/Z_{\mathcal C} + \omega_2(1+x^2) - \omega_1 xy\, , \nonumber \\
 v & = & -y\nu_3/Z_{\mathcal C} + \omega_1(1-y^2) + \omega_2 xy\, .
\label{Eq46}
\end{eqnarray}

\subsubsection{One parameter motion models}

A very simple motion model, considering high-speed scenarios, assumes very
small rotations, such that they can be neglected $\mathbf{R} \approx \mathbf{I}$.
Hence, the rotation matrix approximately equals the identity. In addition it is assumed
that much larger longitudinal $\nu_3$ than horizontal $\nu_2$ and vertical $\nu_1$ movements exist, 
thus the lateral and transversal components of the translation approximately equal zero $\nu_1, \nu_2 \approx 0$. 
This pure 1D translational model is for example used for outlier rejection in highway scenarios \cite{Buczko2016}:
\begin{eqnarray}
 u & = & -x\nu_3/Z_{\mathcal C} \, , \nonumber \\
 v & = & -y\nu_3/Z_{\mathcal C} \, . \nonumber
\end{eqnarray}

\begin{figure}
\begin{centering}
\includegraphics[width=0.6\linewidth]{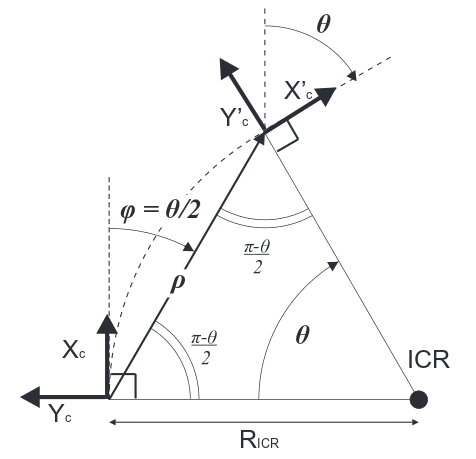}
\caption{Circular motion model of Scaramuzza \cite{Scaramuzza2009}.}
\label{figCicularMotion}
\end{centering}
\end{figure}

The most advanced one parameter motion model is the one introduced by Scaramuzza \cite{Scaramuzza2009}, the so called circular motion model.
It assumes locally planar motion $\nu_1, \omega_2, \omega_3 \approx 0$, so there is only translational velocity $\nu_2, \nu_3$ in the plane parallel to the camera frame
and yaw $\omega_1$. Under planar motion, the two relative poses of a camera
can be described by two parameters, namely the yaw angle
$\theta$ and the polar coordinates $(\rho ,\varphi)$ of the second position
relative to the first position as can be seen in figure \ref{figCicularMotion}. 
Since when using only one camera the scale factor is unknown, we can arbitrarily
set $\rho$ at 1. From this it follows that only two parameters need
to be estimated and so only two image points are required.
However, if the camera moves locally along a circumference
and the x-axis of the camera is set perpendicular
to the radius $R_{ICR}$, then we have $\varphi=\theta/2$; 
thus, only $\theta$ needs to be estimated and so only one image point is required.
Observe that straight motion is also described through the circular motion model; 
in fact in this case we would have $\theta = 0$ and thus $\varphi = 0$.
Comparing the derivation in \cite{Scaramuzza2009} for the discrete motion case with the continuous motion case, we get
$\nu_2 \approx -\sin(\theta/2)$, $\nu_3 \approx \cos(\theta/2)$ and $\omega_1 \approx \cos(\theta)$ and arrive at 
\eqref{Eq8} with the following re-parametrization:
\begin{eqnarray}
 u & = & -x\cos(\theta/2)\underbrace{\rho/Z_{\mathcal C}}_{unknown} - \cos(\theta) xy\, , \nonumber \\
 v & = & (-\sin(\theta/2)-y\cos(\theta/2))\underbrace{\rho/Z_{\mathcal C}}_{unknown} + \cos(\theta)(1-y^2)\, . \nonumber
\end{eqnarray}
In \cite{Scaramuzza2009} there is an efficient way to solve the epipolar constraint under the circular motion model with the \textit{1-point-RANSAC} method.

\section{Ego-motion Estimation Techniques}

\subsection{The Indirect Method}

\begin{figure}
\begin{centering}
\includegraphics[width=0.6\linewidth]{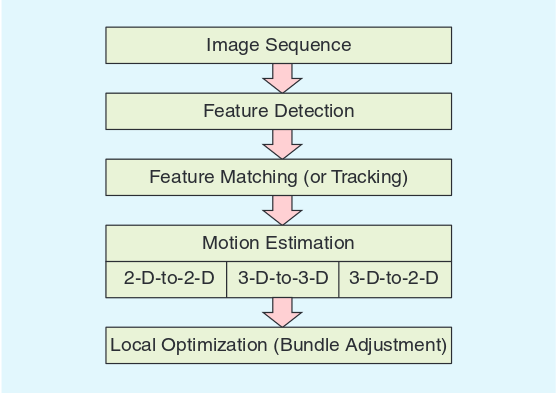}
\caption{The classical visual odometry pipeline taken from Scaramuzza \cite{Scaramuzza2011}.}
\label{figVOPipeline}
\end{centering}
\end{figure}

The indirect method is the most often used classical visual odometry pipeline \cite{Scaramuzza2011} as shown in figure \ref{figVOPipeline}.
It consists of a purely feedforward system architecture, first applying some feature detection, afterwards applying some feature
matching, then estimating the ego-motion and finally refine the ego-motion estimate using some variant of local bundle adjustment.
Here, the brightness constancy assumption is used for the feature matching and afterwards the epipolar constraint is used for the ego-motion estimation.

\subsection{The Direct Method}

Instead of deviding the two optimization problems for feature matching and ego-motion estimation, each driven by a different constraint alone, namely the
photometric and the epipolar constraint, the direct method combines the two constraints into one objective function and directly solves for the ego-motion
without explicitly matching features. In other words, the feature matching is constraint not only by the brightness constancy assumption but also by the
epipolar constraint that reduces the space of possible feature matches to the subspace that is reachable by some ego-motion.
Starting from the Lucas-Kanade objective function \eqref{Eq32}
\begin{equation}
 J(\mathbf{p}) = \sum\limits_{\mathbf{x} \in \mathcal{N}_\mathbf{x}} \left[I_2(\mathcal{T}(\mathbf{x};\mathbf{p}))-I_1(\mathbf{x})\right]^2\,,
\end{equation}
we define the reprojected coordinates after some ego-motion already formulated in \eqref{Eq29} for the reprojection error as a warp
\begin{equation}
 \mathcal{T}(\mathbf{x};\mathbf{p}):= \mathcal{T}(\mathbf{x};\Delta\mathbf{R};\Delta\mathbf{T};Z_{\mathcal C}(\mathbf{x})) = \pi(\Delta\mathbf{R}Z_{\mathcal C}(\mathbf{x})\mathbf{x}+\Delta\mathbf{T})\, ,
\end{equation}
and plug this warp into the Lucas-Kanade objective
\begin{equation}
 J(\Delta\mathbf{R},\Delta\mathbf{T}) = \sum\limits_{\mathbf{x} \in \mathcal{F}_\mathbf{x}} \left[I_2(\mathcal{T}(\mathbf{x};\Delta\mathbf{R};\Delta\mathbf{T};Z_{\mathcal C}(\mathbf{x})))-I_1(\mathbf{x})\right]^2\,.
\label{Eq49}
\end{equation}
Now, this new objective is minimized for the ego-motion $(\Delta\mathbf{R},\Delta\mathbf{T})$
\begin{equation}
 \widehat{\Delta\mathbf{R}},\widehat{\Delta\mathbf{T}} = \mbox{argmin}_{\Delta\mathbf{R},\Delta\mathbf{T}} J(\Delta\mathbf{R},\Delta\mathbf{T})\,,
\end{equation}
instead of the flow $\mathbf{p}$ (which corresponds to the feature matches).
As a result, the formulation in one objective function is just the Lucas-Kanade objective function with a warp that is given by the reprojected coordinates after some ego-motion
and applied to the whole feature set $\mathcal{F}_{\mathbf{x}}$ (sparse approach) or to all pixels $\mathbf{x}$ in the image (dense approach).

This approach was used by a lot of researchers, like \cite{Forster2014, Engel2015} and many others. Note, that for the general case, the depth $Z_{\mathcal C}(\mathbf{x})$
needs to be known for optimization.
A more constraint variation of the direct method is used by Stein et al. \cite{Stein2000}. Here, a reduced motion model (see section \ref{SecMotion}) and the planar ground assumption
(see section \ref{SecStructure}) is integrated to reduce the parameter space and get rid of the depth.


\subsection{Direct versus Indirect Method}

To answer the question which technique should be used cannot clearly be answered because it depends on the
application you would like to realize and what constraints you have in terms of computational effort and accuracy.
Best results can be found by careful intertwining the different optimization objectives of the direct and indirect method.
Having a look at the currently best visual odometry approaches in the Kitti benchmark \cite{Forster2014, Cvisic2015, Buczko2016b}
it turns out that the direct method is best for a fast and robust outlier rejection plus a good initialization of the ego-motion.
For refinement and most accurate estimates the indirect method including bundle adjustment outperforms the direct method.

In general, the non-convex optimization problem of visual odometry is solved best via linearization and iterative least squares including both 
a careful initialization of the motion hypothesis and a 
careful feature selection integrated in a stepwise optimization procedure that adds different constraints and relaxation of constraints in an alternating fashion.
The secret lies in the careful intertwining of all those ingredients. So the classical visual odometry pipeline is extended to have several feedback loops along
the feed forward path \cite{Buczko2018}. 

\subsection{Monocular Visual Odometry: How to get the unknown scale?}

\begin{figure}
\begin{centering}
\includegraphics[width=0.6\linewidth]{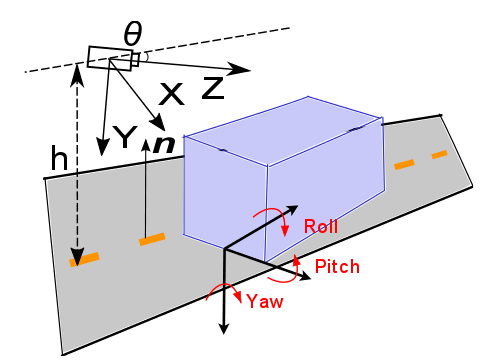}
\caption{The height over ground and planar surface model for scale estimation including additional references from moving obstacles taken from \cite{Song2014}.}
\label{figStreetCamera}
\end{centering}
\end{figure}

\begin{figure}
\begin{centering}
\includegraphics[width=0.6\linewidth]{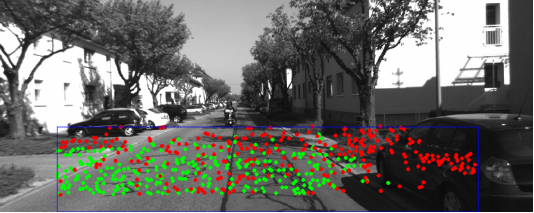}
\caption{Ground plane estimation from 3D points lying on the ground (green points) using the homography constraint taken from \cite{Grater2015}.}
\label{figScale1}
\end{centering}
\end{figure}

\begin{figure}
\begin{centering}
\includegraphics[width=0.6\linewidth]{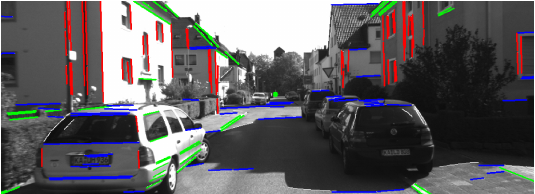}
\caption{Ground plane normal estimation from vanishing point using vertical lines extraction (red lines) that are assumed to be perpendicular to the ground plane taken from \cite{Grater2015}.}
\label{figScale2}
\end{centering}
\end{figure}

As seen so far and derived in section \ref{SecScale}, the motion of a car can be estimated from only one camera up to an unknown scale $\gamma$.
So the only difference between mono and stereo odometry is the estimation of the scale. In stereo systems the scale can directly be estimated from
measurements of 3D point coordinates. In monocular systems the scale has to be estimated via additional apriori knowledge of the scene.
That means, each stereo visual odometry and all the different methods that are based on 3D coordinates can also be used by a monocular system if the scale is estimated in parallel.
Once the scale is known,  the 3D coordinates can be reconstruced using scene reconstruction algorithms like sketched in section \ref{SecScale}.
To summaries, scale drift correction is an integral component of monocular visual odometry. In practice,
it is the single most important aspect that ensures accuracy \cite{Song2014}.
For automotive applications the most prominent way to estimate this unknown scale is to have an apriori model of the structure of the environment as pointed out in
section \ref{SecStructure} and an additional absolute reference value between at least two 3D points in the scene. The apriori knowledge about an absolute reference can
be for example the width or length of a car or a traffic sign \cite{Song2014}. The most often used reference is the height $h$ above ground of the camera mounted on the car
and the most often used scene model is a planar surface of the ground \cite{Stein2000, Geiger2011a, Grater2015}, like visualized in figure \ref{figStreetCamera}. 
In \cite{Grater2015} two methods are combined, in order to obtain the orientation of the ground plane:
First, fitting a plane to the reconstructed point cloud (see figure \ref{figScale1}) and second, deducing the plane normal from vanishing points in the image of a calibrated
camera (see figure \ref{figScale2}).
These methods are complementary. In rural areas, where the vanishing directions are
challenging to be calculated correctly, the plane computation from reconstructed points
leads to an accurate plane estimate. In contrast, the vanishing point estimation performs
best in urban areas, where the scene structure is more dense and the ground plane is
more likely to be occluded.

\section{Outlier Rejection}
\label{SecOutlier}
The essential part of any visual odometry system is the detection of outliers. Therefore, a broad variety of methods has been introduced:
Purely \textbf{flow-based} approaches can be found in \cite{rejectionviarotation, videobasedoutlierremoval,opticflowoutliers}.
All of them are based on the assumption, that the flow follows patterns which are induced by the egomotion of the car. 
Next, \textbf{motion model-based} approaches for outlier detection exist, that explicitly constrain the flow using a certain motion model as in \cite{Scaramuzza2009}.

The majority of existing systems use \textbf{reprojection error-based} approaches.
Here mainly two different ways for finding a proper inlier set are used.
The first one is \textbf{RANSAC} \cite{ransacOriginal}, which is based on the following principle:
In each iteration, a minimum number of random samples is taken from the correspondences to create a motion hypothesis.
Then, a score for each feature is calculated that describes whether it supports the hypothesis.
If the motion estimate reaches a predefined support of the features, the non-supporting features are marked as outliers.
Otherwise, a new random sample is drawn and the next iteration starts.
In order to define the support of a feature in this RANSAC-scheme, the authors of
\cite{robustselectivestereoslam, 
visualodometrybasedonstereoimagesequences, twoyearsofvisualodometry, oneptransacsfm} 
calculate the reprojection error for each feature and compare it to a constant threshold.
Trying to optimize the random process of finding the right hypothesis to separate the features into inliers and outliers,
numerous extensions were created. 
A comparison between the most prominent ones can be found in \cite{ransacComparison}.
\\Due to the random selection of correspondences one can not expect a steady improvement of the resulting motion estimation during the iterations.
Coping with this problem, an alternative method was applied in \cite{aheadwearableshortbaselinestereosystem,robuststereoegomotion,visualodometryusingsparsebundleadjustment}.
Following the naming that was used for RANSAC this class of methods can be united under the notation \textbf{MAximum Subset Outlier Removal} (MASOR).
Here the maximum number of features instead of a minimum random sample is taken to calculate a motion hypothesis. 
This motion estimation and a subsequent outlier rejection step are repeated in an iterative scheme.
Then a support score is calculated for every feature.
Instead of judging the hypothesis, the score is interpreted as a measure for the quality of each feature, as the hypothesis is considered to be a good estimate.
Non-supporting features are rejected and the next iteration starts with the remaining features. 
The process is repeated until a termination criterion is met.
This approach is a good alternative to RANSAC if the number of inliers is sufficient
enough to create a hypothesis that is good enough to separate the outliers.

Following the classical visual odometry pipeline, it can be assumed that for each point $p_i$ the depth $\lambda_i^t \in \mathbb{R}$ is measured by some stereo vision algorithm, 
the image coordinates $\mathbf{x}_i^{t-1}$ are extracted by some feature detector and the correspondent image coordinates in the next frame $\mathbf{x}_i^t$ are measured 
by some optical flow algorithm. To find the optimal estimate of the pose change $(\hat{\mathbf{R}}, \hat{\mathbf{T}})$ very precisely some bundle adjustment approach
minimizing reprojection errors with an iterative gradient descent method has to be carried out like explained in section \ref{SecBundle}.
Here, some carefully chosen initial guess for the pose change has to be given.

The problem of outlier rejection can be formulated as follows: Given the set of all extracted features, we need to find suitable features -- the inliers -- and 
reject all other features from the set -- the outliers. This is usually done by selecting only features with reliable measurements $\{\lambda_i^t, \mathbf{x}_i^{t-1},\mathbf{x}_i^t\}$
and defining some criterion to evaluate how well these measurements fit to some hypothesis of the estimate $(\tilde{\mathbf{R}}, \tilde{\mathbf{T}})$.

\subsection{Reliability of a measurement}

In general it is already known and stated in 
\cite{Song2014} that high translational errors occur at large longitudinal pose changes along the optical axis. The translation estimates 
get especially poor for long distance features \cite{Persson2015}.

Thinking about reliability taking into account the principles of projection, the image resolution, basics of epipolar geometry and the umbiguity of patch matching, 
the reliability of a measurement has two aspects. 

First, since in most of the visual odometry systems depth $\lambda_i^t = b/d_i^t$ is reconstructed from disparity $d_i^t$ 
using a stereo rig with a fixed known baseline $b$ and both the disparity $d_i^t$ and the pairs $\{\mathbf{x}_i^{t-1}, \mathbf{x}_i^t\}$ are based 
on a correspondence search that is done with some optical flow algorithm, for both types of correspondences (within and across time) 
only unambiguous correspondences, e.g. not facing the aperture problem, should be taken into account.

Second, the accuracy of these correspondences are limited by the resolution of the images. 
So even if the correspondences are unambiguous the smaller their distances 
$|\!|\mathbf{x}_i^t-\mathbf{x}_i^{t-1}|\!|$ and $d_i^t$, the less accurate the pose change can be estimated. This is because the ratios 
$|\!|\mathbf{x}_i^t-\mathbf{x}_i^{t-1}|\!| / \Delta p$ and $d_i^t / \Delta p$ between distances 
$|\!|\mathbf{x}_i^t-\mathbf{x}_i^{t-1}|\!|$, $d_i^t$ and the limited image resolution $\Delta p$ are getting smaller with smaller distances and 
thus the signal-to-resolution-ratio decreases. 
Especially for the accuracy of the reconstructed depth $\lambda_i^t = b/d_i^t$, this is crucial because the resolution of depth $\partial\lambda_i^t \propto \partial d_i^t (\lambda_i^t)^2$
reduces quadratically with distance.

Considering these facts, it seems to be easy to figure out good features. Choose near features with large optical flow that are based on highly confident correspondence estimates.

From an algorithmic point of few, there are some further techniques to rely on in order to reduce the number of unreliable measurements that check the measurements for consistency.
Three different consistency checks can befound in the literature: First, the \textit{forward-backward} consistency check for optical flow correspondences. 
Second, the \textit{left-right} consistency check
for stereo-vision corresponences. Third, so called \textit{circular matching} is the combination of \textit{forward-backward} and \textit{left-right} in a circle around a set of four images
of two consecutive stereo image pairs.

Additionally, each correspondence has to fulfill the epipolar constraint for one optimal estimate $(\hat{\mathbf{R}}, \hat{\mathbf{T}})$,
thus the features have to be projections of static points in the scene only. 
Since we cannot guarantee that the measurements are all confident and we do not have the 
optimal pose change estimate at hand, we need to find a good hypothesis $(\tilde{\mathbf{R}}, \tilde{\mathbf{T}})$ and a proper criterion to keep as much suitable features as possible.

\subsection{Using Only Optical Flow}

If only the optical flow is used to detect outliers a certain assuption about the spatial configuration of the flow field has to be given.
The most general assumption is local continuity of the flow field which means that the gradient in the flow field does not exceed some certain threshold.
Here, the threshold separates between a smooth and a discontinuous flow.
Most often used continuity assumptions are parameterized motion models like locally affine flow fields which is true for front-to-parallel moving planes or locally homographic flow fields
which is true for any moving plane.

\subsection{Using Optical Flow and Motion Model}

As already pointed out in \ref{SecOutlier} there are two different categories to recursively get a robust motion hypothesis
by alternating between motion hypothesis estimation based on a fixed feature set and feature set optimization based on a fixed motion hypothesis.
The RANSAC method is very well known and different ways how to realise can be found in the literature (see also section \ref{SecOutlier}). 
The MASOR approaches do not rely on a random sample but on a fixed deterministic feature set that is iteratively reduced to some minimal set.
The two most important variations are given in the following. Starting from a fixed set $F^t_p=\{f_i^t\}_{i=1}^{N_p}$ at iteration number $p=0$, in each
iteration $p$ the motion hypothesis $\hat{\mathbf{R}}_p, \hat{\mathbf{T}}_p$ is refined using the reprojection error for all current inliers 
and afterwards the feature set is updated using some adaptive threshold.

In 2005, the authors of \cite{visualodometryusingsparsebundleadjustment} applied the following criterion to classify outliers: 
\begin{equation}
f_i^t 
\begin{cases}
\in \mathcal F^t_p , & \mbox{if}  \quad \epsilon(i,t,\hat{\mathbf{R}}_p, \hat{\mathbf{T}}_p)- \mu_p < 1.5 \sigma_p\,, \\ 
                                 \notin \mathcal F^t_p, & \mbox{else} \,.
\end{cases}
\label{Eq_1_repro_outlier}
\end{equation}
With mean error $\mu_p=1/N_p \sum_i^{N_p} \epsilon_i^t\left(\hat{\mathbf{R}}_p, \hat{\mathbf{T}}_p\right)$ and squared standard deviation $\sigma_p^2 = 1/N_p \sum_i^{N_p} \left(\epsilon_i^t\left(\hat{\mathbf{R}}_p, \hat{\mathbf{T}}_p\right)-\mu_p\right)^2$. 
The total number of iterations was set to a fixed value.
In 2011, the authors of \cite{aheadwearableshortbaselinestereosystem} changed the criterion slightly:
\begin{equation}
f_i^t 
\begin{cases}
\in \mathcal F^t_p , & \mbox{if}  \quad \epsilon(i,t,\hat{\mathbf{R}}_p, \hat{\mathbf{T}}_p) < 3^2 \mu_p\,, \\ 
                                 \notin \mathcal F^t_p, & \mbox{else} \,.      
\end{cases}
\label{Eq_1_repro_outlier2}
\end{equation}

\subsection{Using Optical Flow and Constraint Scene \& Motion Models}

In order to reduce the space of possible inliers even more, scene knowledge and reduced motion models can be applied to get a more robust motion estimation
that is more robust against outliers.
A very efficient way is the constraint direct method published in \cite{Stein2000}. Here, a reduced motion model (see section \ref{SecMotion}) and the planar ground assumption
(see section \ref{SecStructure}) is integrated to reduce the parameter space, get independent on depth measurements and robustify the ego-motion estimate.

Starting from equation \eqref{Eq46} which is a reduced motion model
\begin{eqnarray}
 u & = & -x\nu_3/Z_{\mathcal C} + \omega_2(1+x^2) - \omega_1 xy\, , \nonumber \\
 v & = & -y\nu_3/Z_{\mathcal C} + \omega_1(1-y^2) + \omega_2 xy\, , \nonumber
\end{eqnarray}
and replace the inverse of the depth by the ground plane assumption
\begin{equation}
 \frac{1}{Z_{\mathcal C}} = \frac{n_1}{d}x+\frac{n_2}{d}y+\frac{n_3}{d}\, ,
\end{equation}
leads to the reduced ground-plane-motion model
\begin{eqnarray}
 u & = & -(\frac{n_1}{d}x+\frac{n_2}{d}y+\frac{n_3}{d})x\nu_3 + \omega_2(1+x^2) - \omega_1 xy\, , \nonumber \\
 v & = & -(\frac{n_1}{d}x+\frac{n_2}{d}y+\frac{n_3}{d})y\nu_3 + \omega_1(1-y^2) + \omega_2 xy\, . \nonumber
\end{eqnarray}
If one assumes that the ground plane is parallel to the $YZ$-plane, then $n_1=n_3=0$, $n_2=1$ and the ground-plane-motion model
reduces to
\begin{eqnarray}
 u & = & -\underbrace{\frac{\nu_3}{d}}_{a}yx + \omega_2(1+x^2) - \omega_1 xy\, , \nonumber \\
 v & = & -\underbrace{\frac{\nu_3}{d}}_{a}y^2 + \omega_1(1-y^2) + \omega_2 xy\, . \nonumber
\end{eqnarray}
This model has to be plugged into the objective \eqref{Eq49} of the direct method and can then be optimized for
$a$, $\omega_1$ and $\omega_2$. If the height over ground which now equals $d$ is known, then $a$ can be solved for $\nu_3=ad$.

In \cite{Klappstein2006} an additional constraint is used to detect moving objects exploiting the available constraint envelope of a 3D point.
The epipolar constraint expresses that the viewing rays of a static 3D point (the lines joining the projection centers and the 3D point)
must meet. A moving 3D point in general induces skew viewing rays violating the constraint.

\begin{figure}
\begin{centering}
\includegraphics[width=0.6\linewidth]{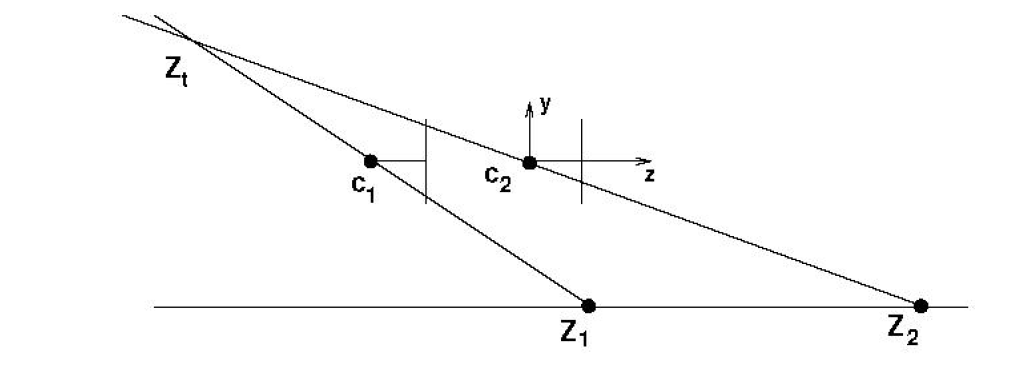}
\caption{Side view: Positive depth constraint. The camera is moving
from $c_l$ to $c_2$. A 3D point on the road is moving from $Z_1$ to $Z_2$
which is a greater distance than the camera's movement (overtaking
vehicle). The triangulated 3D point $Z_t$ lies behind the camera,
violating the constraint. This example is taken from \cite{Klappstein2006}.}
\label{figPDepth}
\end{centering}
\end{figure}

The fact that all points seen by the camera must lie in front of it is known as the positive depth constraint.
See figure \ref{figPDepth} for details.
This constraint is independent of the scene structure.
In order to apply the constraint the translation direction
(forward or backward) of the camera has to be known
(in addition to the essential matrix).
If points intersect behind the camera,
the 3D point itself must be moving.

\begin{figure}
\begin{centering}
\includegraphics[width=0.6\linewidth]{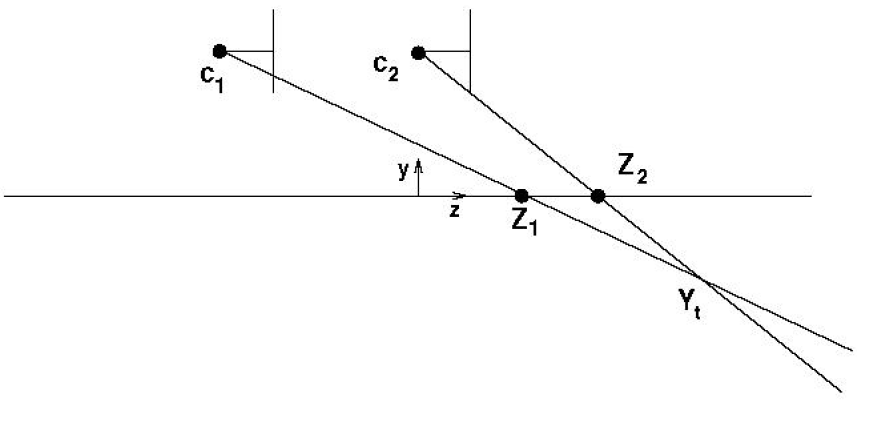}
\caption{Side view: Positive height constraint. The camera is
moving from $c_l$ to $c_2$. A 3D point on the road is moving from
$Z_1$ to $Z_2$ which is a smaller distance than the camera's movement
(vehicle ahead). The triangulated 3D point $Y_t$ lies under the road,
violating the constraint. This example is taken from \cite{Klappstein2006}.}
\label{figPHeight}
\end{centering}
\end{figure}

Traffic driving in front of the ego-vehicle
with lower or identical speed
(preceding traffic) is detected
by the positive height constraint. For details see figure \ref{figPHeight}.
All 3D points must lie above the road.
This constraint is not as
powerful as the positive depth constraint
since it applies only for image points
under the horizon.
Furthermore the geometry of the road has to be known.

\subsection{Explicit Segmentation of Multiple Objects}

Instead of just finding outliers that do not fit to some ego-motion hypothesis and classifying these outliers as individually moving objects the
multiple ego-motions of all of the traffic participants could in principle explicitly be estimated in parallel using the 
multibody epipolar constraint for the discrete motion case \cite{Ma2004}. The minimal setting is the existence of two independent moving objects with motions $(\mathbf{R}_1,\mathbf{T}_1)$ and
 $(\mathbf{R}_2,\mathbf{T}_2)$. This leads to the following multibody epipolar constraint:
\begin{equation}
 \left(\overline{\mathbf{x}}^T(t+\Delta t)\widehat{\Delta\mathbf{T}}_1\Delta\mathbf{R}_1\overline{\mathbf{x}}(t) \right)\left(\overline{\mathbf{x}}^T(t+\Delta t)\widehat{\Delta\mathbf{T}}_2\Delta\mathbf{R}_2\overline{\mathbf{x}}(t)\right)= 0\, .
\end{equation}
If there are only two independent moving objects (or at least enough features for both motions that are far more than the number of outliers) then both ego-motions 
can be recovered with an algorithm. For more then two rigid-body motions it is getting hard to solve but is still possible under some constraints \cite{Ma2004}.
Here, we have another problem, that is to find the correct number of multiple motions. This is an additional segmentation problem that has to be solved in advance or in conjunction
with the estimation of the multiple motions. Several approaches do already exist that use this basic idea, for example \cite{Schindler2014}. Here for example simplifications are introduced and solutions are 
provided to circumvent the segmentation problem e.g. using the planar epipolar constraint for a large number of fixed small motion patches \cite{Vogel2013, Vogel2015}.
To segment moving objects that are non-rigid which are composed of rigid moving parts with arbitrary shape learning techniques (preferably unsupervised) can be applied to learn specific motion patterns 
related to certain movement classes \cite{Guthier2012, Guthier2013, Guthier2015}. 

\subsection{Robust estimators}

There is an alterative (or additional) method to handle outliers. Instead of rejection via some consistency check, one could use robust estimators that weaken the
influence of an outlier within the objective function \cite{Scaramuzza2011}. 
Therefore, the squared error for the estimates of the unknowns has to be replaced by a different robust error norm, like the infinity-norm (the absolute values)
to set up objective functions. The advantage of such an approach is that no additional rejection mechanism has to be set up to reject outliers. The drawback of robust estimators
is, that the outliers still do influence the estimation result and the minimization of robust objective functions is computationally much harder than simple squared error norms.
Of course, robust estimators can be applied in conjunction with an outlier rejection mechanism.


\begin{thebibliography}{1}


\bibitem{Ma2004} 
    Ma, Y., Soatto, S., Kosecka, J., Sastry, S. S. (2012): 
    An invitation to 3-d vision: from images to geometric models. 
    Springer Science \& Business Media.

\bibitem{Hartley2003}
    Hartley, R.,  Zisserman, A. (2003):
    Multiple view geometry in computer vision. 
    Cambridge university press.

\bibitem{Willert2008a} 
    Willert, V., Schmuedderich, J., Eggert, J., Goerick, C., Koerner, E. (2008): 
    Probabilistic Optical Flow Estimation for Large Pixel Displacements Utilizing Egomotion Flow Compensation. 
    In British Machine Vision Conference (BMVC) (pp. 1-10).

\bibitem{Schmuedderich2008}
   Schmuedderich, J., Willert, V., Eggert, J., Rebhan, S., Goerick, C., Sagerer, G., Koerner, E. (2008):
   Estimating Objects Proper Motion using Optical Flow, Kinematics and Depth Information.
   In: IEEE Transactions on Systems, Man and Cybernetics, Part B, 38 (4) 1139 - 1151.

\bibitem{Willert2009b}
   Willert, V., Schmuedderich, J., Rebhan, S., Eggert, J. (2009):
   Estimating objects proper motion using optical flow, kinematics, and depth information.
   European Patent.

\bibitem{Higgins1980} 
    Longuet-Higgins, H. C., Prazdny, K. (1980): 
    The interpretation of a moving retinal image. 
    Proceedings of the Royal Society of London B: Biological Sciences, 208(1173), 385-397.

\bibitem{Willert2006} 
    Willert, V. (2006): 
    Recurrent visual motion segmentation. 
    PhD Thesis, TU Darmstadt, VDI Fortschritt-Berichte, Reihe 8, Nr. 1101.

\bibitem{Toussaint2007}
   Toussaint, M., Willert, V., Eggert, J., Koerner, E. (2007): 
   Motion Segmentation Using Inference in Dynamic Bayesian Networks. 
   In BMVC (pp. 1-10).

\bibitem{Willert2012} 
    Willert, V. (2012): 
    Image processing for engineers -- Basics of image-guided measuring and control. 
    Lecture slides, TU Darmstadt.

\bibitem{Buczko2016} 
    Buczko, M., Willert, V. (2016):
    How to Distinguish Inliers from Outliers in Visual Odometry for High-Speed Automotive Applications.
    In Intelligent Vehicles Symposium (IV), 2016 IEEE (pp. ). IEEE.

\bibitem{Buczko2016b} 
   Buczko, M.,  Willert, V. (2016): 
   Flow-decoupled normalized reprojection error for visual odometry. 
   In 2016 IEEE 19th International Conference on Intelligent Transportation Systems (ITSC) (pp. 1161-1167). IEEE.

\bibitem{Buczko2017} 
   Buczko, M., Willert, V. (2017): 
   Monocular outlier detection for visual odometry. 
   In 2017 IEEE Intelligent Vehicles Symposium (IV) (pp. 739-745). IEEE.

\bibitem{Nister2004}
   Nistér, D. (2004): 
   An efficient solution to the five-point relative pose problem. 
   Pattern Analysis and Machine Intelligence, IEEE Transactions on, 26(6), 756-770.

 \bibitem{Hartley2012}
    Hartley, R., Li, H. (2012): 
    An efficient hidden variable approach to minimal-case camera motion estimation. 
    Pattern Analysis and Machine Intelligence, IEEE Transactions on, 34(12), 2303-2314.

\bibitem{Scaramuzza2011}
   Scaramuzza, D., Fraundorfer, F. (2011): 
   Visual odometry: Part I: The first 30 years and fundamentals. 
   IEEE Robotics and Automation Magazine, 18(4), 80-92.

\bibitem{Fraundorfer2012}
   Fraundorfer, F., Scaramuzza, D. (2012): 
   Visual odometry: Part II: Matching, robustness, optimization, and applications. 
   Robotics and Automation Magazine, IEEE, 19(2), 78-90.

\bibitem{Thormaehlen2004}
   Thormaehlen, T., Broszio, H., Weissenfeld, A. (2004): 
   Keyframe selection for camera motion and structure estimation from multiple views. 
   In Computer Vision-ECCV 2004 (pp. 523-535). Springer Berlin Heidelberg.

\bibitem{Forster2014}
    Forster, C., Pizzoli, M., Scaramuzza, D. (2014): 
    SVO: Fast semi-direct monocular visual odometry. 
    In Robotics and Automation (ICRA), 2014 IEEE International Conference on (pp. 15-22). IEEE.

\bibitem{Engel2015} 
     Engel, J., Stückler, J., Cremers, D. (2015): 
     Large-scale direct slam with stereo cameras. 
     In Proceedings of the IEEE International Conference on Intelligent Robots and Systems (IROS).

\bibitem{Kanade1981}
  Lucas, B. D., Kanade, T. (1981): 
  An iterative image registration technique with an application to stereo vision. 
  In IJCAI (Vol. 81, pp. 674-679).

 \bibitem{Bouguet2001}
  Bouguet, J. Y. (2001): 
  Pyramidal implementation of the affine lucas kanade feature tracker description of the algorithm. 
  Intel Corporation, 5(1-10), 4.

\bibitem{Baker2004}
  Baker, S., Matthews, I. (2004): 
  Lucas-Kanade 20 years on: A unifying framework. 
  International journal of computer vision, 56(3), 221-255.

\bibitem{Willert2007}
  Willert, V., Toussaint, M., Eggert, J., Koerner, E. (2007): 
  Uncertainty optimization for robust dynamic optical flow estimation. 
  In Sixth International Conference on Machine Learning and Applications (ICMLA 2007) (pp. 450-457). IEEE.

\bibitem{Willert2008b}
  Willert, V., Eggert, J., Toussaint, M., Koerner, E. (2008):
  Probabilistic Exploitation of the Lucas and Kanade Smoothness Constraint. 
  In Machine Learning and Applications, 2008. ICMLA'08. Seventh International Conference on (pp. 259-266). IEEE.

\bibitem{Willert2005} 
    Willert, V., Eggert, J., Adamy, J.,  Koerner, E. (2005):
    Non-gaussian velocity distributions integrated over space, time, and scales. 
    IEEE Transactions on Systems, Man, and Cybernetics, Part B (Cybernetics), 36(3), 482-493.

\bibitem{Willert2009a}
  Willert, V., Eggert, J. (2009): 
  A stochastic dynamical system for optical flow estimation. 
  In Computer Vision Workshops (ICCV Workshops), 2009 IEEE 12th International Conference on (pp. 711-718). IEEE.

\bibitem{Baker2003}
  Baker, S., Matthews, I. (2003): 
  Lucas-Kanade 20 years on: A unifying framework: Part 2. 
  The Robotics Institute, Carnegie Mellon University.

\bibitem{Buczko2018} 
   Buczko, M., Willert, V., Schwehr, J., Adamy, J. (2018): 
   Self-validation for automotive visual odometry. 
   In 2018 IEEE Intelligent Vehicles Symposium (IV) (pp. 1-6). IEEE.

\bibitem{Schreier2014}
    Schreier, M. Willert, V., Adamy, J. (2014):
    Grid Mapping in Dynamic Road Environments: Classification of Dynamic Cell Hypothesis via Tracking.
    In: IEEE International Conference on Robotics and Automation (ICRA), Hong Kong, China.

\bibitem{Schreier2012}
    Schreier, M.,  Willert, V. (2012): 
    Robust free space detection in occupancy grid maps by methods of image analysis and dynamic B-spline contour tracking. 
    In 2012 15th International IEEE Conference on Intelligent Transportation Systems (pp. 514-521). IEEE.

\bibitem{Schreier2013}
    Schreier, M., Willert, V.,  Adamy, J. (2013): 
    From grid maps to parametric free space maps?A highly compact, generic environment representation for ADAS. 
    In 2013 IEEE Intelligent Vehicles Symposium (IV) (pp. 938-944). IEEE.

\bibitem{Schreier2015}
    Schreier, M., Willert, V.,  Adamy, J. (2015): 
    Compact representation of dynamic driving environments for ADAS by parametric free space and dynamic object maps. 
    IEEE Transactions on Intelligent Transportation Systems, 17(2), 367-384.

\bibitem{Wedel2009}
  Wedel, A., Badino, H., Rabe, C., Loose, H., Franke, U., Cremers, D. (2009): 
  B-spline modeling of road surfaces with an application to free-space estimation. 
  Intelligent Transportation Systems, IEEE Transactions on, 10(4), 572-583.

\bibitem{Neumann2015}
  Neumann, L., Vanholme, B., Gressmann, M., Bachmann, A., Kahlke, L., Schule, F. (2015): 
  Free Space Detection: A Corner Stone of Automated Driving. 
  In Intelligent Transportation Systems (ITSC), 2015 IEEE 18th International Conference on (pp. 1280-1285). IEEE.

\bibitem{Stein2000} 
    Stein, G. P., Mano, O., Shashua, A. (2000): 
    A robust method for computing vehicle ego-motion. 
    In Intelligent Vehicles Symposium, 2000. IV 2000. Proceedings of the IEEE (pp. 362-368). IEEE.

\bibitem{Scaramuzza2009}
  Scaramuzza, D., Fraundorfer, F., Siegwart, R. (2009): 
  Real-time monocular visual odometry for on-road vehicles with 1-point ransac. 
  In Robotics and Automation, 2009. ICRA'09. IEEE International Conference on (pp. 4293-4299). IEEE.

\bibitem{Engel2014}
    Engel, J., Schoeps, T., Cremers, D. (2014): 
    LSD-SLAM: Large-scale direct monocular SLAM. 
    In Computer Vision?ECCV 2014 (pp. 834-849). Springer International Publishing.

\bibitem{Cvisic2015} 
    Cvisic, I., Petrovic, I. (2015):
    Stereo odometry based on careful feature selection and tracking.
    In European Conference on Mobile Robots 2015.

\bibitem{Song2014}
  Song, S., Chandraker, M. (2014): 
  Robust scale estimation in real-time monocular SFM for autonomous driving. 
  In Computer Vision and Pattern Recognition (CVPR), 2014 IEEE Conference on (pp. 1566-1573). IEEE.

\bibitem{Geiger2011a}
  Geiger, A., Ziegler, J., Stiller, C. (2011):
  Stereoscan: Dense 3d reconstruction in real-time. 
  IEEE Intelligent Vehicles Symposium (Baden-Baden, Germany).

\bibitem{Grater2015}
   Grater, J., Schwarze, T., Lauer, M. (2015). 
   Robust scale estimation for monocular visual odometry using structure from motion and vanishing points. 
   In Intelligent Vehicles Symposium (IV), 2015 IEEE (pp. 475-480). IEEE.

\bibitem{rejectionviarotation}
  A.~Amit and E.~Rivlin and I.~Shimshoni (2001):
  ROR: Rejection of outliers by rotations.
  IEEE Transactions on Pattern Analysis and Machine Intelligence.

\bibitem{videobasedoutlierremoval}
  B.~Grinstead and others (2006):
  Improving video-based robot self localization through outlier removal.
  Proceedings of the 1st Joint Emer. Prep. \& Response/Robotic \& Remote Sys. Top. Mtg.

\bibitem{opticflowoutliers}
  P.~Santana and L.~Correia (2008):
  Improving visual odometry by removing outliers in optic flow.

\bibitem{robuststereoegomotion}
  C.F.~Olson and L.H.~Matthies and M.~Schoppers and M.W.~Maimone (2000):
  Robust Stereo Ego-motion for Long Distance Navigation
  IEEE Conference on Computer Vision and Pattern Recognition. Proceedings.

\bibitem{visualodometryusingsparsebundleadjustment}
  N.~S\"underhauf and K.~Konolige and S.~Lacroix and P.~Protzel (2005):
  Visual Odometry Using Sparse Bundle Adjustment on an Autonomous Outdoor Vehicle.
  Autonome Mobile Systeme 2005,Springer Berlin Heidelberg.

\bibitem{aheadwearableshortbaselinestereosystem}
  H.~Badino and T.~Kanade (2011):
  A Head-Wearable Short-Baseline Stereo System for the Simultaneous Estimation of Structure and Motion.
  IAPR Conference on Machine Vision Application.

\bibitem{ransacOriginal}
  M.~Fischler and R.~Bolles (1981):
  Random Sample Consensus: A Paradigm for Model Fitting with Applications to Image Analysis and Automated Cartography.
  Communications of the ACM 24.6.

\bibitem{twoyearsofvisualodometry}
  M.~Maimone and C.~Yang and L.~Matthies (2007):
  Two years of visual odometry on the mars exploration rovers.
  Journal of Field Robotics 24.3.

\bibitem{visualodometrybasedonstereoimagesequences}
  B.~Kitt and A.~Geiger and H. Lategahn (2010):
  Visual odometry based on stereo image sequences with ransac-based outlier rejection scheme.
  IEEE Intelligent Vehicles Symposium.

\bibitem{robustselectivestereoslam}
  F.~Bellavia and others (2013):
  Robust selective stereo slam without loop closure and bundle adjustment.
  International Conference on Image Analysis and Processing, Springer Berlin Heidelberg.

\bibitem{oneptransacsfm}
  D.~Scaramuzza (2011):
  1-point-ransac structure from motion for vehicle-mounted cameras by exploiting non-holonomic constraints.
  International Journal of Computer Vision 95.1.

\bibitem{ransacComparison}
  R.~Raguram and F.~M.~Frahm and M.~Pollefeys (2008):
  A comparative analysis of RANSAC techniques leading to adaptive real-time random sample consensus.
  European Conference on Computer Vision.

\bibitem{Persson2015} 
    Persson, M., Piccini, T., Felsberg, M., Mester, R. (2015):
    Robust stereo visual odometry from monocular techniques.
    In Intelligent Vehicles Symposium (IV), 2015 IEEE (pp. 686-691). IEEE.

\bibitem{Klappstein2006} 
   Klappstein, J., Stein, F., Franke, U. (2006): 
   Monocular motion detection using spatial constraints in a unified manner. 
   In Intelligent Vehicles Symposium, 2006 IEEE (pp. 261-267). IEEE.

\bibitem{Schindler2014}
  Schindler, K., Suter, D., Wang, H. (2008): 
  A model-selection framework for multibody structure-and-motion of image sequences. 
  International Journal of Computer Vision, 79(2), 159-177.

\bibitem{Vogel2013}
    Vogel, C., Schindler, K., Roth, S. (2013): 
    Piecewise rigid scene flow. 
    In Computer Vision (ICCV), 2013 IEEE International Conference on (pp. 1377-1384). IEEE.

\bibitem{Vogel2015}
    Vogel, C., Schindler, K., Roth, S. (2015): 
    3D Scene Flow Estimation with a Piecewise Rigid Scene Model. 
    International Journal of Computer Vision, 115(1), 1-28.

\bibitem{Guthier2012}
    Guthier, T., Eggert, J., Willert, V. (2012): 
    Unsupervised learning of motion patterns. 
    In ESANN.

\bibitem{Guthier2013}
    Guthier, T., Willert, V., Schnall, A., Kreuter, K., Eggert, J. (2013): 
    Non-negative sparse coding for motion extraction. 
    In The 2013 International Joint conference on neural networks (IJCNN) (pp. 1-8). IEEE.

\bibitem{Guthier2015}
   Guthier, T., Willert, V., Eggert, J. (2015): 
   Topological sparse learning of dynamic form patterns. 
   Neural computation, 27(1), 42-73.

\end{thebibliography}
\end{document}